\documentclass[letterpaper]{article} 
\usepackage[preprint]{aaai2027}

\usepackage[hyphens]{url}  
\usepackage{graphicx} 
\usepackage{natbib}  
\usepackage{caption} 
\usepackage{multirow}
\usepackage{algorithm}
\usepackage{algorithmic}

\usepackage{graphicx}
\usepackage{booktabs}
\usepackage{multirow}
\usepackage{svg}
\usepackage{newfloat}
\usepackage{listings}
\DeclareCaptionStyle{ruled}{labelfont=normalfont,labelsep=colon,strut=off} 
\floatstyle{ruled}
\newfloat{listing}{tb}{lst}{}
\floatname{listing}{Listing}
\usepackage{graphicx}
\usepackage{tabularx}
\usepackage{array}

\usepackage{booktabs}
\usepackage{xspace}
\usepackage{amsmath}
\usepackage{amssymb}
\title{ChorusTIC: Training-Free Multivariate Time Series Classification \\ via Chorus In-Context Learning}
\author{
    Juntao Fang\textsuperscript{\rm 1}\thanks{This work was done during an internship at Huawei Noah's Ark Lab.},
    Shifeng Xie\textsuperscript{\rm 2,\rm 3},
    Ruichu Cai\textsuperscript{\rm 1}\corresponding,
    Shengji Zheng\textsuperscript{\rm 1},
    Zijian Li\textsuperscript{\rm 4},
    Keli Zhang\textsuperscript{\rm 2},
    Lujia Pan\textsuperscript{\rm 2},
    Themis Palpanas\textsuperscript{\rm 3},
    Zhifeng Hao\textsuperscript{\rm 5}
}

\affiliations{
    \textsuperscript{\rm 1}Guangdong University of Technology\\
    \textsuperscript{\rm 2}Huawei Noah's Ark Lab\\
    \textsuperscript{\rm 3}Universit\'e Paris Cit\'e\\
    \textsuperscript{\rm 4}Mohamed bin Zayed University of Artificial Intelligence\\
    \textsuperscript{\rm 5}Shantou University\\
    Corresponding author: cairuichu@gmail.com
}

\begin{document}

\maketitle


\begin{abstract}


Time series classification underpins applications in healthcare, sensing, and industrial monitoring. Although time series foundation models have advanced forecasting and transferable representation learning, classification still typically requires training a task-specific classifier for each target dataset. Moreover, multivariate time series are often processed channel by channel, limiting the modeling of dependencies among variables. We introduce ChorusTIC, a foundation model for in-context time series classification across heterogeneous channel configurations without target-task parameter updates. ChorusTIC combines Random Subchannel Slot Concatenation, applied consistently within each episode, with a shared dual-axis encoder to jointly model temporal dynamics and interactions across channels. It maps varying channel configurations into a fixed-width representation independent of the original channel count, calibrates feature axes using context-derived distributions, and predicts query labels through in-context learning. ChorusTIC is pretrained solely on synthetic labeled episodes in which context and query samples share a task background and classes are distinguished by sparse temporal patterns or interactions across channels. Evaluations on the complete UEA-30 and UCR-128 archives show strong performance with both full context and limited labeled data, without fitting a classifier for each target dataset. Code is available at \url{https://github.com/DMIRLAB-Group/ChorusTIC}.

\end{abstract}


\section{Introduction}
Time series classification (TSC) supports applications including human activity recognition, clinical monitoring, digital health, and industrial sensing~\cite{TSClassificationSurvey,uea,convtran}. Many such applications involve multivariate time series recorded simultaneously by multiple sensors or electrodes ~\cite{uea,convtran}. Discriminative evidence may be localized to particular variables and temporal intervals~\cite{laxcat,shapenet}, while multivariate classification may also depend on interactions among variables, correlations across sensors, and temporal lead and lag relationships~\cite{leadlag, SVPT,GAC,MPTSNet}. A transferable multivariate classifier must therefore capture both within-channel temporal patterns and task-relevant cross-channel relationships while accommodating heterogeneous channel configurations.

Recent time series foundation models (TSFMs) have demonstrated
promising transferability across datasets and
domains~\cite{TSFMsurvey}. For classification, however, the prevailing
approach remains representation transfer: a pretrained encoder produces
features for each sample, after which a task-specific classifier is
fitted on every target
dataset~\cite{moment,mantis,mantisv2,nutime,unishape}. Although pretraining provides reusable representations across tasks,
this pipeline still requires target-task optimization and remains
sensitive to the choice of representation layer, token aggregation
strategy, and downstream classifier~\cite{TIC-FM}.
Moreover, multivariate inputs are often processed through channel-wise
encoding, which may not preserve task-relevant temporal and
cross-channel interactions. In-context learning (ICL) provides an
alternative by conditioning predictions directly on labeled
examples~\cite{TIC-FM,tict,TimeEE,rocketpfn}. Given a labeled context
set and an unlabeled query set, an in-context classifier infers the
target decision rule without updating its parameters. Existing
approaches, however, focus primarily on univariate TSC and do not
jointly address two challenges in multivariate classification across
heterogeneous channel configurations: modeling aligned temporal and
cross-channel interactions and mapping variable channel sets to a
fixed-dimensional representation for support-conditioned inference.

We propose ChorusTIC, a classification-native foundation model that
performs Chorus ICL across heterogeneous channel configurations. At
the signal level, Random Subchannel Slot Concatenation (RSSC) assigns
input channels to episode-consistent group-slot positions. A shared
dual-axis encoder captures temporal and within-group cross-channel
interactions, after which fixed-order slot concatenation produces a
representation whose dimensionality is independent of the original
channel count. At the task level, Column Distribution Modeling
calibrates the resulting feature axes using the labeled context, and
row-wise interaction forms sample-level representations. A
leakage-protected in-context classifier injects labels only into
context representations and predicts query labels without target-task
parameter updates.

Training this model requires pretraining tasks that capture the relationship between context and query samples rather than collections of isolated sequences. Existing episodic generators are primarily designed for univariate classification or derive class identity from
a restricted family of generative mechanisms~\cite{tict,TimeEE}. We therefore construct a labeled multivariate episodic prior. Each episode shares a task-level temporal background, while classes differ through sparse temporal or cross-channel rules applied to selected temporal regions and channel subsets. The rule families cover temporal motifs, position and order changes, informative-channel selection, cross-channel phase and delay relationships, and correlation changes. Instance-level nuisance transformations increase within-class diversity, while episode-wise label permutation prevents fixed associations between synthetic patterns and numerical label indices. 

Our contributions are summarized as follows:
\begin{itemize}
    \item We introduce ChorusTIC, a classification-native foundation
    model that performs support-conditioned inference across
    heterogeneous univariate and multivariate classification tasks
    without target-task parameter updates.


    \item We construct a labeled multivariate episodic prior whose
    classes differ through sparse temporal and cross-channel
    discriminative rules under a shared task background, together
    with instance-level variation and episode-wise label permutation.

    \item We evaluate ChorusTIC under full-context and low-label
    protocols on complete UEA-30 and UCR-128 archives,
    together with ablation studies that assess its architectural, inference, and pretraining designs.
\end{itemize}

\section{Related Work}

\paragraph{Time series foundation models (TSFMs).}
Time series forecasting  represents one of the most active areas of foundation model research. Large pretrained forecasting models support a range of deployment protocols, including zero-shot prediction, adaptation from limited observations, and task-specific fine-tuning \citep{chronos,ansari2025chronos2,timesfm11,toto,moirai,auer:25tirex,tempopfn,lagllama}. By contrast, foundation models for time series classification commonly follow a representation-transfer paradigm: a pretrained encoder extracts features, and a separate classifier is then fitted using labeled samples from each target dataset \citep{mantis,nutime,timesbert,cauker,auer:25tirexclassification,units,gpt2TS}. General-purpose models such as MOMENT \citep{moment} also adopt this embedding-based formulation and are widely used as representation backbones for downstream classification. Therefore, although existing methods provide transferable time series representations, broadly applicable frameworks that directly infer query labels for unseen classification tasks without target-specific optimization remain limited. 

\paragraph{In-context time series classification.}
In-context learning predicts query labels from labeled context examples without fitting a task-specific classifier. TIC-FM ~\cite{TIC-FM} combines a pretrained time series encoder with a latent-memory in-context learner, while TiCT~\cite{tict} is trained end to end on synthetic episodes and introduces scalable label representations. Both are primarily developed or evaluated for univariate classification. TableTime~\cite{tabletime} and FETA~\cite{feta} use general-purpose language models with textual tables or channel-wise exemplar reasoning, whereas iAmTime~\cite{iamtime} treats classification as one task within a general instruction-conditioned framework. Concurrently, RocketPFN~\cite{rocketpfn} combines random convolutional features with a pretrained tabular in-context classifier. TimEE ~\cite{TimeEE} constructs augmented classification tasks from the training splits of UCR datasets to train an in-context classifier. These methods demonstrate the potential of training-free time series
classification but do not jointly learn classification-specific temporal and cross-channel representations while accommodating variable channel counts. ChorusTIC addresses this gap through dual-axis encoding and episode-consistent fixed-dimensional composition. 

\paragraph{Synthetic pretraining.}
Synthetic data support the construction of forecasting corpora, representation-learning datasets, and complete classification tasks~\cite{chronos,cauker,mantisv2,tict,TimeEE}. ChorusTIC complements these efforts with a multivariate episodic prior aligned with its deployment protocol: context and query samples share a task-level background, while sparse temporal or cross-channel discriminative rules determine class identity. Further discussion
and detailed comparisons appear in Appendix A.

\section{Method}
\label{sec:method}

\subsection{Problem Formulation and Model Overview}
\label{subsec:overview}

For a classification task $\tau$, let
$\mathcal{C}_{\tau}=\{(X_i^c,y_i^c)\}_{i=1}^{N_c}$ and
$\mathcal{Q}_{\tau}=\{X_j^q\}_{j=1}^{N_q}$ denote the labeled
context and unlabeled query sets, respectively. Each
$X_i^c,X_j^q\in\mathbb{R}^{C\times L}$ contains $C$ channels and
$L$ time steps, with task-specific $C$ and $L$ fixed within $\tau$.
Let $\mathbf{X}^c=[X_1^c;\ldots;X_{N_c}^c]$ and
$\mathbf{X}^q=[X_1^q;\ldots;X_{N_q}^q]$ be the stacked inputs, and
let $Y^c=(y_1^c,\ldots,y_{N_c}^c)$ and
$Y^q=(y_1^q,\ldots,y_{N_q}^q)$ be their labels. We model
$p_{\Theta}(Y^q\mid\mathbf{X}^q,\mathcal{C}_{\tau})$ without
target-task parameter updates.

\begin{figure*}[t]
    \centering
    \includegraphics[width=0.98\textwidth]{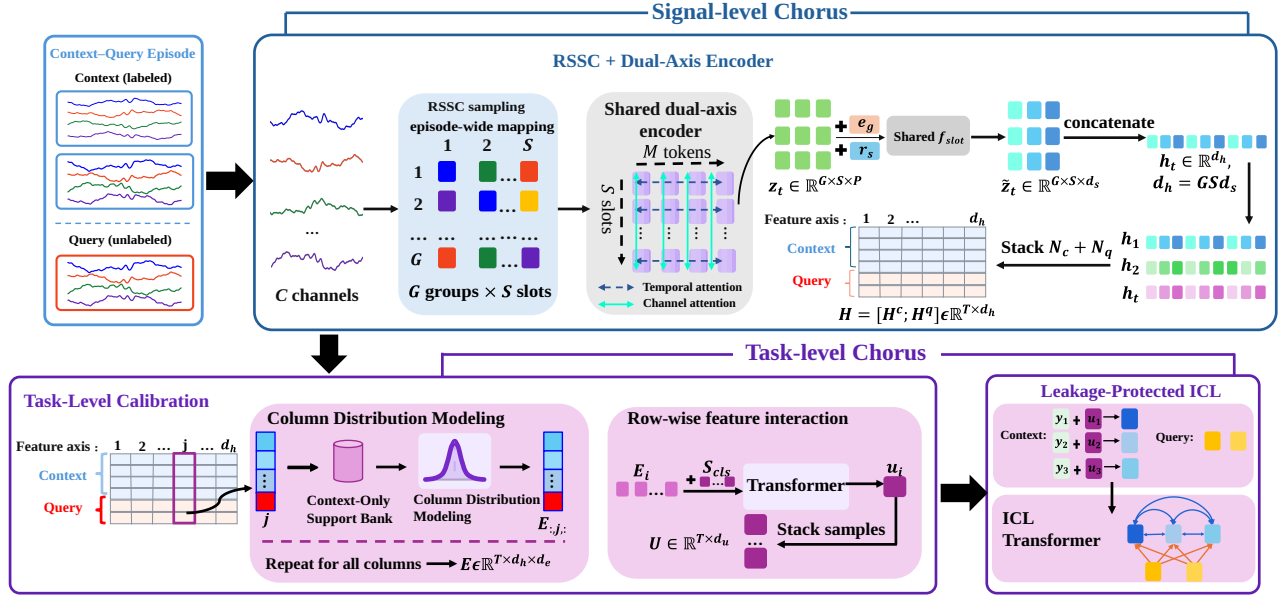}
    \caption{
        \textbf{Overview of ChorusTIC.}
Given labeled context and unlabeled queries, the signal-level Chorus
uses one RSSC channel-to-slot assignment throughout the episode. Each
sampled group is processed by a shared dual-axis encoder that captures
temporal structure within slots and cross-channel interactions across
slots. A shared readout summarizes each encoded slot, and fixed-order
concatenation yields a fixed-width representation for each sample. The
task-level Chorus calibrates feature axes from context-only
distributions before row-wise interaction. Finally, the
leakage-protected ICL Transformer predicts each query from the labeled
context while preventing direct information exchange between queries.
    }
    \label{fig:ChorusTIC}
\end{figure*}

As illustrated in Figure~\ref{fig:ChorusTIC}, ChorusTIC performs
Chorus ICL in three stages. First, RSSC samples an episode-level
channel-to-slot assignment, and a shared dual-axis encoder models
temporal and within-group cross-channel interactions before
fixed-order slot composition. Second, Column Distribution Modeling
calibrates feature axes using context-derived distributions, and
row-wise interaction forms sample-level representations. Third, a
leakage-protected ICL Transformer conditions on context labels and
predicts all queries in parallel.

Let $T=N_c+N_q$,
$\mathbf{X}_{\tau}=[\mathbf{X}^c;\mathbf{X}^q]
\in\mathbb{R}^{T\times C\times L}$, and let $\mathcal{I}$ denote
the RSSC assignment shared across the episode. The overall computation
is
\begin{equation*}
\begin{aligned}
H &= \mathcal{R}_{\psi}(\mathbf{X}_{\tau};\mathcal{I}),
\qquad
U = \mathcal{A}_{\phi}(H;N_c),\\
O^q &= \mathcal{G}_{\theta}(U,Y^c;N_c).
\end{aligned}
\label{eq:overall_pipeline}
\end{equation*}
Here, $H\in\mathbb{R}^{T\times d_h}$,
$U\in\mathbb{R}^{T\times d_u}$, and
$O^q\in\mathbb{R}^{N_q\times K}$.
The operators $\mathcal{R}_{\psi}$, $\mathcal{A}_{\phi}$, and
$\mathcal{G}_{\theta}$ denote the RSSC-based signal encoder,
task-level calibration and row-wise interaction, and the
leakage-protected in-context classifier, respectively.
Moreover, $Y^c\in\{1,\ldots,K\}^{N_c}$ and $K$ is the number of
classes in the current episode. We consider $K\leq K_{\max}$ in the
main text; Appendix~B covers $K>K_{\max}$. Predictions are obtained
as $P^q=\operatorname{softmax}(O^q)$ and
$\hat{y}_j^q=\arg\max_{1\leq k\leq K}P_{j,k}^q$ for
$j=1,\ldots,N_q$.

\subsection{RSSC-Based Multivariate Representation}
\label{subsec:rssc_encoder}

\paragraph{RSSC definition.}
Random Subchannel Slot Concatenation is an episode-level 
adapter that maps a variable-size channel set to a representation
whose width is independent of the original channel count. RSSC
consists of two operations surrounding a shared group encoder:
(i) an episode-consistent assignment from input channels to ordered
group-slot positions, and (ii) a fixed-order composition of the
encoded slot representations. The dual-axis encoder is the shared
group encoder applied between these two RSSC operations.

Unless otherwise stated, we use coverage sampling. When $C\geq N_s$,
RSSC samples $N_s$ channels without replacement. When $C<N_s$,
independently permuted channel lists are concatenated until all slots
are filled:
\begin{equation*}
\mathbf{i}
=
\begin{cases}
\operatorname{Perm}(\mathcal{V})_{1:N_s},
& C\geq N_s,\\[1.5mm]
\bigl[
\operatorname{Perm}_1(\mathcal{V});
\ldots;
\operatorname{Perm}_{R}(\mathcal{V})
\bigr]_{1:N_s},
& C<N_s,
\end{cases}
\label{eq:rssc_coverage_sampling}
\end{equation*}
where $R=\lceil N_s/C\rceil$. Thus, observed channels are reused
when necessary rather than replaced by artificial zero-valued slots.
The index vector is reshaped into $G$ ordered groups
\begin{equation*}
\mathcal{I}_g
=
(i_{g,1},\ldots,i_{g,S})
\in\mathcal{V}^{S},
\qquad
g=1,\ldots,G.
\label{eq:channel_groups}
\end{equation*}
The assignment
$\mathcal{I}=\{\mathcal{I}_g\}_{g=1}^{G}$ is shared across all context
and query samples in an episode. Thus, each group-slot position $(g,s)$
identifies a fixed source channel within the episode, although the
assignment may change across episodes. For sample $t$, the input to
group $g$ is
$X_{t,g}=X_{t,\mathcal{I}_g}\in\mathbb{R}^{S\times L}$.

Each group defines a sampled subchannel view, and channel-axis
attention operates only among its $S$ slots. Thus, $S$ controls the
number of channels modeled jointly within each group, whereas $G$
controls the number of sampled views. The groups are not jointly
processed by channel-axis attention; instead, their encoded slots are
concatenated and subsequently integrated by the row-wise Transformer.

\paragraph{Patch tokenization and shared dual-axis encoder.}
Each selected channel is resampled to length $L_0$ and divided into
$M$ non-overlapping patches. Each patch is encoded from its normalized
values, first differences, and local statistics:
$u_{t,g,s,m}
=
f_{\mathrm{tok}}(x_{t,g,s,m},
\Delta x_{t,g,s,m},
\mu_{t,g,s,m},
\sigma_{t,g,s,m})
\in\mathbb{R}^{P}$.
Stacking the tokens within a group gives
$U_{t,g}^{(0)}\in\mathbb{R}^{S\times M\times P}$.
More details are provided in Appendix~B.

Each dual-axis layer first models temporal dependencies within each
slot and then interactions across slots at aligned patch positions:
\begin{equation*}
\begin{aligned}
\widetilde{U}_{t,g}^{(\ell)}
&=
\operatorname{TempBlock}^{(\ell)}
\left(U_{t,g}^{(\ell)}\right),\\
U_{t,g}^{(\ell+1)}
&=
\operatorname{ChanBlock}^{(\ell)}
\left(\widetilde{U}_{t,g}^{(\ell)}\right),
\end{aligned}
\quad
\ell=0,\ldots,L_D-1.
\label{eq:dual_axis}
\end{equation*}
The temporal block attends over the $M$ patches independently for
each slot, whereas the channel block attends over the $S$ slots
independently at each aligned patch position. A shared summary-token
readout then produces an encoded slot representation $z_{t,g,s}\in\mathbb{R}^{P}$ for every group-slot position. Because channel-axis attention has
already mixed information among the sampled slots,
$z_{t,g,s}$ is conditioned on the other channels in group $g$ and
is not an independently encoded channel representation.

\paragraph{Fixed-dimensional RSSC composition.}
Learnable group embeddings $e_g\in\mathbb{R}^{P}$ and slot
embeddings $r_s\in\mathbb{R}^{P}$ encode structural positions in
the fixed group-slot layout rather than globally aligned sensor
identities. Each encoded slot is projected as
\begin{equation*}
\widetilde{z}_{t,g,s}
=
f_{\mathrm{slot}}
\left(z_{t,g,s}+e_g+r_s\right)
\in\mathbb{R}^{d_s},
\label{eq:rssc_slot_projection}
\end{equation*}
where $f_{\mathrm{slot}}:\mathbb{R}^{P}\rightarrow
\mathbb{R}^{d_s}$ is shared across all group-slot positions. The sample representation
is obtained by concatenating the projected slots in a fixed order:
\begin{equation*}
h_t
=
\operatorname{LN}\!\left(
\mathop{\Vert}_{g=1}^{G}
\mathop{\Vert}_{s=1}^{S}
\widetilde{z}_{t,g,s}
\right)
\in\mathbb{R}^{d_h},
\qquad
d_h=GSd_s.
\label{eq:rssc_composition}
\end{equation*}
For fixed $G$, $S$, and $d_s$, the representation dimension
$d_h$ is independent of the original channel count $C$. RSSC therefore retains the positions of multiple sampled subchannel views while providing a common representation width across tasks. It does not assume globally aligned channel identities: each slot has consistent task-local semantics within an episode but may be reassigned across episodes. Stacking all context and query representations gives
\begin{equation*}
H
=
[H^c;H^q]
=
[h_1;\ldots;h_T]
\in\mathbb{R}^{T\times d_h}.
\label{eq:episode_matrix}
\end{equation*}

\subsection{Task-Level Calibration and ICL}
\label{subsec:task_inference}

\paragraph{Column Distribution Modeling.}
The semantics and scale of a latent feature axis may vary across
tasks. For each feature axis $j\in\{1,\ldots,d_h\}$, we first embed
every scalar activation:
\begin{equation*}
q_{i,j}^{(0)}
=
f_{\mathrm{cell}}(H_{i,j})
\in\mathbb{R}^{d_e},
\qquad
Q_j^{(0)}
=
[q_{1,j}^{(0)};\ldots;q_{T,j}^{(0)}].
\label{eq:cell_embedding}
\end{equation*}
A shared induced-attention encoder constructs an axis-specific bank
from context cells only:
\begin{equation*}
S_j
=
\mathcal{S}_{\phi}
\!\left(Q_{j,1:N_c}^{(0)}\right)
\in\mathbb{R}^{N_{\mathrm{ind}}\times d_e},
\label{eq:support_bank}
\end{equation*}
where $N_{\mathrm{ind}}$ denotes the number of inducing tokens.
Every context or query cell attends to this bank:
\begin{equation*}
R_j
=
\mathcal{C}_{\phi}(Q_j^{(0)},S_j),
\qquad
[W_j,B_j]
=
\mathcal{D}_{\phi}(R_j),
\label{eq:column_model}
\end{equation*}
where $W_j,B_j\in\mathbb{R}^{T\times d_e}$. The calibrated
embeddings are
\begin{equation*}
E_{:,j,:}
=
W_j\odot
\left(H_{:,j}\mathbf{1}_{d_e}^{\top}\right)+B_j.
\label{eq:column_calibration}
\end{equation*}
Applying this operation to all $d_h$ axes yields
$E\in\mathbb{R}^{T\times d_h\times d_e}$.

Because $S_j$ is constructed exclusively from context samples, query samples cannot modify the reference distribution or communicate with one another. Column Distribution Modeling processes each feature axis independently, while cross-axis interactions are handled by the subsequent row-wise Transformer.

\paragraph{Row-wise feature interaction.}
For sample $i$, let $E_i=E_{i,:,:}\in\mathbb{R}^{d_h\times d_e}$. We prepend $K_{\mathrm{cls}}$ learnable summary tokens $S_{\mathrm{cls}}\in\mathbb{R}^{K_{\mathrm{cls}}\times d_e}$ and obtain the sample representation as $u_i = \operatorname{vec}\!\left( \mathcal{F}_{\mathrm{row}} ([S_{\mathrm{cls}};E_i])_{1:K_{\mathrm{cls}},:} \right) \in\mathbb{R}^{d_u}$, where $d_u=K_{\mathrm{cls}}d_e$. The row Transformer processes each sample independently, preventing cross-sample information flow. Stacking the outputs gives $U=[u_1;\ldots;u_T]\in\mathbb{R}^{T\times d_u}$.

\paragraph{Leakage-protected ICL.}
Labels are injected only into context tokens:
\begin{equation}
\bar{u}_i=
\begin{cases}
u_i+\mathcal{E}_y(y_i^c), & i\leq N_c,\\
u_i, & i>N_c,
\end{cases}
\label{eq:label_injection}
\end{equation}
where $\mathcal{E}_y(\cdot)\in\mathbb{R}^{d_u}$ is a learnable label embedding. Let $\bar{U}=[\bar{u}_1;\ldots;\bar{u}_T]$. The ICL Transformer uses context tokens as its only keys and
values. With $i$ denoting the target position and $j$ the source
position, the additive attention mask is
\begin{equation}
M_{ij}
=
\begin{cases}
0, & j\leq N_c,\\
-\infty, & j>N_c.
\end{cases}
\label{eq:context_only_mask}
\end{equation}
Thus, context tokens attend only to the context, and each query
attends only to the context. A query retains its own representation
through the residual stream but never serves as a key or value,
preventing query-to-query information flow. The query logits are
\begin{equation}
V
=
\mathcal{G}_{\theta}(\bar{U};M),
\qquad
O^q
=
\operatorname{Dec}
\left(V_{N_c+1:T}\right).
\label{eq:query_logits}
\end{equation}

\subsection{Labeled Episodic Pretraining}
\label{subsec:episodic_prior}

ChorusTIC is pretrained on classification episodes rather
than isolated sequences. Each episode samples
$\omega=(K,C,L,N_c,N_q,\kappa,r,d)$, where $\kappa$ denotes the task
type, $r$ the discriminative rule family, and $d$ the task-difficulty
setting. Univariate and multivariate tasks are sampled with
probabilities $0.2$ and $0.8$, respectively. Univariate tasks use $C=1$ and
$r\in\mathcal{R}_{\mathrm{temp}}$, whereas multivariate tasks use
$2\leq C\leq10$ and $r\in\mathcal{R}_{\mathrm{cross}}$. Class
proportions follow
$\boldsymbol{\varrho}\sim\operatorname{Dirichlet}
(\alpha\mathbf{1}_K)$, with every class represented in the context set.


\paragraph{Shared background and discriminative rules.}
Each episode first samples a shared temporal background:
\[b\sim\operatorname{Categorical}(\boldsymbol{\lambda}),
\qquad
W\sim\mathcal{P}_b,
\qquad
W\in\mathbb{R}^{C\times L},\]where $\{\mathcal{P}_b\}$ is a collection of temporal process families. For each class $k$, a sparse rule operator constructs a prototype
\begin{equation*}
P_k =
\Gamma_{r,k}
\left(W;\mathcal{S}_k,\mathcal{T}_k,\eta_k\right),
\qquad
k=1,\ldots,K,
\label{eq:class_prototype}
\end{equation*}
where $\mathcal{S}_k$ and $\mathcal{T}_k$ denote the informative channel subset and temporal region, respectively, and $\eta_k$ contains the rule parameters. Temporal rule families introduce class-dependent motif shape, polarity, position, order, or local anomalies. Cross-channel rule families introduce class differences through informative-channel selection, relative delay or phase, and correlation structure. Each episode uses one sampled discriminative rule family. An instance of class $y_i$ is generated by $X_i = \mathcal{A}_{\xi_i}(P_{y_i})+\varepsilon_i$, where $\mathcal{A}_{\xi_i}$ applies instance-specific nuisance transformations and $\varepsilon_i$ denotes sensor noise. The difficulty variable $d$ controls the discriminative strength and nuisance magnitude. Detailed background families, rule operators, and transformations are provided in Appendix B.

\paragraph{Episode construction and objective.}
Generated samples are divided into context and query sets, with every
query class represented in the context. We then sample an
episode-specific bijection
$\sigma_{\mathcal{E}}:\{1,\ldots,K\}\rightarrow\{1,\ldots,K\}$
and apply it to both context and query labels. Let $\widetilde{\mathcal{C}}_{\mathcal{E}}
= \left\{ (X_i^c,\sigma_{\mathcal{E}}(y_i^c)) \right\}_{i=1}^{N_c}$. The model is trained by minimizing query cross-entropy:
\begin{equation*}
\mathcal{L}(\Theta)
=
-
\mathbb{E}_{\mathcal{E}\sim p_{\mathrm{syn}}}
\left[
\frac{1}{N_q}
\sum_{j=1}^{N_q}
\log
p_{\Theta}
\left(
\sigma_{\mathcal{E}}(y_j^q)
\mid
X_j^q,\widetilde{\mathcal{C}}_{\mathcal{E}}
\right)
\right].
\label{eq:pretraining_objective}
\end{equation*}
Label permutation prevents fixed synthetic rules from acquiring fixed
numerical label meanings and forces the model to infer label semantics
from the context.

\subsection{Deployment-Time Inference}
\label{subsec:deployment}

All parameters remain fixed on a target task. To reduce sensitivity
to arbitrary label indices, we average predictions over
$M_{\pi}$ cyclic label permutations. For
$m=0,\ldots,M_{\pi}-1$, define
\begin{equation*}
\pi_m(y)
=
1+\big((y-1+m)\bmod K\big).
\label{eq:cyclic_label_permutation}
\end{equation*}
Let $P_m\in\{0,1\}^{K\times K}$ be the corresponding permutation
matrix, with $(P_m)_{y,\pi_m(y)}=1$.

Because RSSC samples channel-to-slot assignments stochastically, we additionally average predictions over $M_R$ independent RSSC draws. Within each draw, the same assignment is shared by all context and query samples. Let
\begin{equation*}
O^{q,(m,a)}
=
\operatorname{ChorusTIC}
\left(
X^c,
\pi_m(Y^c),
X^q;
\mathcal{I}^{(a)}
\right)
\end{equation*}
denote the query logits under label permutation $m$ and RSSC draw
$a$. The aligned ensemble logits are
\begin{equation*}
\bar{O}^{q}
=
\frac{1}{M_{\pi}M_R}
\sum_{m=0}^{M_{\pi}-1}
\sum_{a=1}^{M_R}
O^{q,(m,a)}P_m^{\top}.
\label{eq:joint_ensemble}
\end{equation*}
The final probabilities and predictions are
\[
P^q=\operatorname{softmax}(\bar O^q/\tau),
\qquad
\widehat y_j^q=\arg\max_k P_{j,k}^q,
\]
where the temperature is set to $\tau=0.9$ by default.
We use $M_R=4$ in the main experiments. Full ensemble settings and the
hierarchical extension for tasks with $K>K_{\max}$ are provided in
Appendix~B.


\section{Experiments}
\label{sec:experiments}

Our experiments address four questions: (1) Can ChorusTIC classify univariate and multivariate time series without target-task parameter updates? (2) How does it compare with generic ICL methods and frozen TSFMs that fit target-specific classifiers? (3) How effectively does it infer a task-specific decision rule from limited labeled context? (4) How do its architectural, inference, and pretraining components contribute to performance?

\subsection{Experimental Setup}
\label{subsec:experimental_setup}

\paragraph{Benchmarks.}
We evaluate ChorusTIC on the UEA Multivariate Time Series Classification
Archive~\cite{uea} and the UCR Time Series Classification
Archive~\cite{dau2019ucr}. UEA contains 30 multivariate datasets with diverse
channel counts, sequence lengths, and class structures, and serves as our
primary benchmark for native multivariate classification. UCR contains 128
univariate datasets and evaluates transfer to the single-channel setting. We
use the official train/test splits throughout. For ChorusTIC, the training
split provides the labeled context and the test split constitutes the query
set; no model parameter is updated on a target dataset.

\paragraph{Compared methods.}
We organize the baselines by their target-task adaptation protocol.

\emph{Time series ICL classifiers} predict query labels directly from labeled context examples without target-specific parameter updates. On UCR, we compare with TIC-FM~\cite{TIC-FM} and TiCT~\cite{tict}, which provide the closest protocol match in the univariate setting.

\emph{Generic ICL classifiers} include TabICL~\cite{tabicl} and TabICLv2~\cite{tabiclv2}. For each dataset, we concatenate the channel-wise sequences of a sample into a fixed-dimensional vector and treat the resulting samples as rows of a tabular classification task. These methods provide training-free controls, but do not explicitly encode temporal order.

\emph{Frozen time series foundation models} include MOMENT~\cite{moment},
Mantis~\cite{mantis}, MantisV2~\cite{mantisv2},
UniShape~\cite{unishape}, and NuTime~\cite{nutime}. We keep each pretrained
backbone fixed and extract its final-layer representation using the model's
default readout. A lightweight classifier is then fitted on the target
training split. We follow the original frozen-feature protocol when one is
available. Because NuTime is evaluated primarily through fine-tuning, we fit a
random forest to its frozen CLS representations for the main comparison.


\paragraph{Evaluation protocol.}
Across all settings, the complete official test split serves as the
query set. Full-context evaluation uses the complete training split as
labeled context, with no parameter updates to ChorusTIC. For fixed-shot
evaluation, we sample $k\in\{5,10\}$ examples per class and average
results over five context sets shared across methods. A dataset is
excluded at shot level $k$ if any class has fewer than $k$ training
examples; all methods use the same eligible datasets and context sets.
For the context-scaling analysis, we use shared class-stratified
subsets containing $20\%$, $30\%$, $40\%$, $50\%$, or $60\%$ of the
training split.


\paragraph{Metrics and statistical analysis.}
We report the unweighted average classification accuracy across datasets. When describing aggregate gains, relative improvement over a reference method is computed as $(A_{\mathrm{ours}}-A_{\mathrm{ref}})/A_{\mathrm{ref}}\times100\%$ using unrounded average accuracies.


\subsection{Main Results}
\label{subsec:main_results}

\paragraph{Multivariate classification on UEA.}
\label{subsubsec:uea_results}
Table~\ref{tab:uea_main} reports results on the complete UEA-30 archive. ChorusTIC achieves the highest average accuracy and the best average rank among the evaluated methods without target-specific parameter updates. Relative to MantisV2+LR, the strongest frozen-feature baseline, ChorusTIC improves average accuracy by approximately $2.51\%$ and reduces the average rank from $4.02$ to $3.57$. This comparison is notable because MantisV2+LR fits a separate logistic-regression
classifier on every target dataset, whereas ChorusTIC infers the target decision rule directly from labeled context examples.

Among generic ICL baselines, ChorusTIC yields relative improvements of approximately $6.34\%$ over TabICLv2 and $10.62\%$ over TabICL. These methods likewise avoid target-specific fitting but operate on vectorized multivariate inputs without explicit temporal or aligned cross-channel modeling. Their lower aggregate performance is consistent with the benefit of time-series-specific representation
learning for in-context classification.

\begin{table}[t]
  \centering
  \caption{
    \textbf{Classification results on the complete UEA-30 archive.}
    ``Target fit'' indicates whether a dataset-specific classifier is fitted on
    the target training split. Best and second-best average accuracies and
    average ranks are shown in \textbf{bold} and
    \underline{underlined}, respectively. Per-dataset results are provided in Appendix~D.
  }
  \label{tab:uea_main}

  \small
  \setlength{\tabcolsep}{2pt}
  \renewcommand{\arraystretch}{1.05}

  \begin{tabular*}{\columnwidth}{
    @{\extracolsep{\fill}}llccc@{}
  }
    \toprule
    Protocol
      & Method
      & \shortstack{Target\\fit}
      & \shortstack{Avg.\\Acc.}
      & \shortstack{Avg.\\Rank} \\
    \midrule

    Time-series ICL
      & ChorusTIC
      & No
      & \textbf{72.27\%}
      & \textbf{3.57} \\

    \midrule

    \multirow{2}{*}{Generic ICL}
      & TabICL
      & No
      & 65.33\%
      & 5.18 \\

      & TabICLv2
      & No
      & 67.96\%
      & 4.37 \\

    \midrule

    \multirow{6}{*}{Frozen TSFM}
      & MOMENT+SVM
      & Yes
      & 68.17\%
      & 5.48 \\

      & Mantis+RF
      & Yes
      & 69.34\%
      & 5.22 \\

      & MantisV2+LR
      & Yes
      & \underline{70.50\%}
      & \underline{4.02} \\

      & MantisV2+RF
      & Yes
      & 69.54\%
      & 4.63 \\

      & UniShape+RF
      & Yes
      & 69.72\%
      & 4.92 \\

      & NuTime+RF
      & Yes
      & 57.92\%
      & 7.62 \\

    \bottomrule
  \end{tabular*}
\end{table}

\paragraph{Univariate classification on UCR.}
\label{subsubsec:ucr_results}
Table~\ref{tab:ucr_main_1} reports results on the complete UCR-128 archive. ChorusTIC achieves the highest average accuracy and the best average rank among the evaluated methods. Relative to MantisV2+LR, the
strongest frozen-feature baseline, it improves average accuracy by approximately $1.41\%$ while requiring no target-specific classifier. It also reduces the average rank from $5.50$ to $4.43$, indicating consistent performance across the archive.

Among training-free time-series classifiers, ChorusTIC yields relative improvements of approximately $1.44\%$ over TIC-FM and $2.51\%$ over TiCT. It also outperforms TabICLv2, the strongest generic ICL
baseline, by approximately $2.89\%$. These results show that the same pretrained model retains strong performance in the single-channel setting while supporting both univariate and multivariate classification without target-specific optimization.

\begin{table}[t]
  \centering
  \caption{
    \textbf{Classification results on the UCR-128 archive.}
    ``Target fit'' indicates whether a classifier is fitted on the target
    training split. Best and second-best results are shown in
    \textbf{bold} and \underline{underlined}, respectively.
    Per-dataset results are provided in Appendix~D.
  }
  \label{tab:ucr_main_1}

  \small
  \setlength{\tabcolsep}{2pt}
  \renewcommand{\arraystretch}{1.05}

  \begin{tabular*}{\columnwidth}{
    @{\extracolsep{\fill}}llccc@{}
  }
    \toprule
    Protocol
      & Method
      & \shortstack{Target\\fit}
      & \shortstack{Avg.\\Acc.}
      & \shortstack{Avg.\\Rank} \\
    \midrule

    \multirow{6}{*}{Frozen TSFM}
      & MOMENT+SVM
      & Yes
      & 77.98\%
      & 6.11 \\

      & Mantis+RF
      & Yes
      & 78.67\%
      & 6.42 \\

      & MantisV2+RF
      & Yes
      & 78.79\%
      & 6.51 \\

      & MantisV2+LR
      & Yes
      & \underline{80.03\%}
      & 5.50 \\

      & UniShape+RF
      & Yes
      & 78.86\%
      & 5.83 \\

      & NuTime+RF
      & Yes
      & 69.39\%
      & 9.55 \\

    \midrule

    \multirow{2}{*}{Generic ICL}
      & TabICL
      & No
      & 76.83\%
      & 6.38 \\

      & TabICLv2
      & No
      & 78.88\%
      & 5.15 \\

    \midrule

    \multirow{3}{*}{Time-series ICL}
      & TiCT
      & No
      & 79.17\%
      & \underline{4.81} \\

      & TIC-FM
      & No
      & 80.01\%
      & 5.32 \\

      & ChorusTIC
      & No
      & \textbf{81.16\%}
      & \textbf{4.43} \\

    \bottomrule
  \end{tabular*}
\end{table}


\subsection{Low-Label Multivariate Classification}
\label{subsec:low_label}

We examine whether ChorusTIC can infer a target-task decision rule from
limited labeled context. We consider two complementary protocols. In the
fixed-shot protocol, we sample $5$ or $10$ labeled examples per class. In the
proportional protocol, we retain $20\%$--$60\%$ of the official training
split as labeled data. Within each budget, all methods are evaluated on the
same eligible datasets and matched labeled subsets.

\begin{table}[t]
  \centering
  \caption{
  \textbf{Fixed-shot classification accuracy on UEA.}
  Results are averaged over five independently sampled support sets and
  over the 28 and 24 datasets eligible for the 5-shot and 10-shot settings,
  respectively. Within each budget, all methods use the same datasets and
  matched support sets. ``Target fit'' indicates whether a classifier is
  fitted on the target support set. Best and second-best results are shown
  in \textbf{bold} and \underline{underlined}, respectively.
  }
  \label{tab:low_label_main}
  \footnotesize
  \setlength{\tabcolsep}{0pt}
  \renewcommand{\arraystretch}{1.05}
  \begin{tabular*}{\columnwidth}
    {@{\extracolsep{\fill}}lcrr@{}}
    \toprule
    Method & Target fit & 5-shot & 10-shot \\
    \midrule
    MOMENT+SVM
      & Yes
      & 59.42\%
      & 63.64\% \\
    MantisV2+LR
      & Yes
      & 61.50\%
      & \underline{65.80\%} \\
    MantisV2+RF
      & Yes
      & 60.10\%
      & 63.66\% \\
    UniShape+RF
      & Yes
      & \underline{62.19\%}
      & 65.35\% \\
    NuTime+RF
      & Yes
      & 54.68\%
      & 59.83\% \\
    \midrule
    TabICL
      & No
      & 59.48\%
      & 64.06\% \\
    TabICLv2
      & No
      & 58.65\%
      & 61.85\% \\
    \midrule
    ChorusTIC
      & No
      & \textbf{62.88\%}
      & \textbf{68.47\%} \\
    \bottomrule
  \end{tabular*}
\end{table}

\paragraph{Fixed-shot performance.}
Table~\ref{tab:low_label_main} shows that ChorusTIC achieves the highest average accuracy under both label budgets without target-specific parameter updates. With five examples per class, it yields an approximately $1.11\%$ relative improvement over UniShape+RF, the strongest competing method. With ten examples per class, its relative improvement over the strongest baseline, MantisV2+LR, increases to approximately $4.06\%$. 
Compared with TabICL, the strongest generic ICL baseline under both budgets, ChorusTIC yields relative improvements of approximately $5.72\%$ and $6.88\%$ at five and ten shots, respectively. These results indicate that time-series-specific support conditioning enables effective decision-rule inference from limited labeled examples without target-specific optimization.

\begin{figure}[t]
  \centering
  \includegraphics[width=0.95\linewidth]{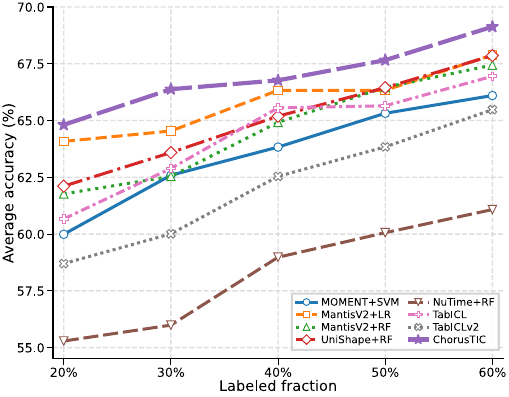}
  \caption{
  \textbf{Scaling with labeled data on UEA-30.}
  Each point reports the average accuracy obtained using the indicated fraction of the official training split.
  }
  \label{fig:low_label_scaling}
\end{figure}

\paragraph{Scaling with labeled context.}
Figure~\ref{fig:low_label_scaling} complements the fixed-shot analysis by
varying the labeled fraction from $20\%$ to $60\%$ on UEA-30. ChorusTIC
ranks first at every reported fraction, yielding relative improvements
of $0.66\%$ to $2.85\%$ over the strongest competing method at each
fraction. Its average accuracy increases monotonically from $64.81\%$
with $20\%$ labeled data to $69.13\%$ with $60\%$, corresponding to a
$6.67\%$ relative increase over its own accuracy at the smallest
reported fraction.

These results indicate that support-conditioned inference remains
effective beyond the fixed-shot regime. As more labeled context becomes
available, ChorusTIC consistently improves without target-specific
parameter updates. Together, the fixed-shot and proportional results
demonstrate its effectiveness across different low-label regimes.

\subsection{Ablation and Pretraining-Prior Analysis}
\label{subsec:ablation}

We evaluate four design choices spanning the architecture, inference
procedure, and pretraining prior: channel-axis attention,
task-conditioned feature calibration, label-permutation ensembling,
and cross-channel discriminative rules. Architecture and prior
ablations are separately pretrained from scratch using the same
optimization schedule, training budget, and random seed as the
complete model. The inference ablation reuses the complete-model
checkpoint and modifies only the label-permutation strategy at test
time. All variants are evaluated under the same unified protocol.

\begin{table}[t]
  \centering
  \caption{
    \textbf{Ablation results on UEA-30.}
    $\Delta$ Acc. is measured relative to the complete model in
    percentage points.
  }
  \label{tab:ablation}

  \small
  \setlength{\tabcolsep}{3pt}
  \renewcommand{\arraystretch}{1.06}

  \begin{tabularx}{\columnwidth}{
    @{}
    >{\raggedright\arraybackslash}p{0.21\columnwidth}
    >{\raggedright\arraybackslash}X
    r
    r
    @{}
  }
    \toprule
    Factor
      & Variant
      & \shortstack{Avg.\\Acc.}
      & \shortstack{$\Delta$\\Acc.} \\
    \midrule

    Complete model
      & ChorusTIC
      & \textbf{72.27\%}
      & 0.00 \\

    \midrule

    \multirow{2}{*}{Architecture}
      & No channel-axis attention
      & 71.69\%
      & $-0.58$ \\

      & No task-conditioned calibration
      & 70.66\%
      & $-1.61$ \\

    \midrule

    Inference
      & No label-permutation ensemble
      & 71.38\%
      & $-0.89$ \\

    \midrule

    Pretraining prior
      & No cross-channel discriminative rules
      & 71.07\%
      & $-1.20$ \\

    \bottomrule
  \end{tabularx}
\end{table}

For the channel-attention ablation, we replace channel-axis attention with an identity mapping while retaining temporal attention. For the calibration ablation, we remove the task-conditioned affine transformation while preserving the representation width and in-context learner. Both variants use the complete episodic prior.
We evaluate label-permutation ensembling by disabling cyclic
permutations while keeping the RSSC ensemble size fixed. For the prior
ablation, we retain multivariate episodes but replace cross-channel
rules involving informative channels, relative phase or delay, and
correlation structure with channel-wise temporal rules.


Table~\ref{tab:ablation} shows that all ablations reduce accuracy,
supporting the complementary roles of the four components.
Task-conditioned calibration has the largest effect, consistent with
the need to align task-dependent RSSC feature axes before in-context
inference. Removing cross-channel rules causes the next-largest
decline, indicating that channel interaction benefits from a prior
that makes cross-channel structure class-discriminative. The
label-permutation result shows that cyclic averaging mitigates
sensitivity to arbitrary label indices. Removing channel-axis
attention lowers accuracy, supporting the benefit of modeling aligned
within-group interactions before slot-level summarization and
task-level integration. Together, these results support the joint use
of signal-level interaction, task-level calibration, inference-time
ensembling, and a matching multivariate episodic prior.

\section{Conclusion}

We introduce ChorusTIC, a classification-native foundation model for support-conditioned univariate and multivariate TSC without per-dataset classifier fitting. ChorusTIC combines episode-consistent RSSC, a shared dual-axis encoder,
context-derived calibration, and leakage-protected ICL to model temporal and cross-channel interactions across heterogeneous channel configurations and predict query labels directly from labeled context.
A labeled episodic prior over synthetic tasks aligns pretraining with deployment. On the UCR-128 and UEA-30 archives, ChorusTIC achieves strong full-context and low-label performance without target-specific classifier fitting and improves consistently as labeled context grows. These results indicate that cross-channel modeling and support-conditioned inference provide complementary mechanisms for classification across heterogeneous channel configurations. Although UCR and UEA provide broad coverage, they do not encompass the full range of deployment conditions. Future work will develop a broader benchmark that extends the current protocols to cover missing channels, asynchronous sampling, and domain shifts.

\bibliography{reference}

\clearpage
\onecolumn
\setcounter{secnumdepth}{3}
\renewcommand{\topfraction}{0.95}
\renewcommand{\bottomfraction}{0.90}
\renewcommand{\textfraction}{0.05}
\renewcommand{\floatpagefraction}{0.80}
\setlength{\textfloatsep}{10pt plus 2pt minus 2pt}
\setlength{\floatsep}{10pt plus 2pt minus 2pt}
\setlength{\intextsep}{10pt plus 2pt minus 2pt}
\numberwithin{equation}{section}
\counterwithin{figure}{section}
\counterwithin{table}{section}
\appendix

\section{Extended Related Work}

\label{app:related_work}

\paragraph{General-purpose and classification-oriented TSFMs.}
Large-scale pretraining enables time series models to transfer temporal
knowledge across datasets and tasks~\citep{TSFMsurvey}. MOMENT adopts
masked time series modeling and evaluates transfer to forecasting,
classification, anomaly detection, and imputation
~\citep{moment}. NuTime decomposes each temporal window into normalized
shape, mean, and standard deviation, and uses numerically multi-scaled
embeddings with contrastive pretraining~\citep{nutime}. Both models
serve primarily as transferable encoders; downstream classification
requires a predictor fitted on the labeled target split.

To improve classification transfer, classification-oriented TSFMs tailor their tokenization schemes and pretraining objectives to
discriminative representation learning.
Mantis introduces a lightweight Transformer with time-series-specific
token generation and contrastive pretraining, together with
multivariate adaptations~\citep{mantis}. MantisV2 and related Mantis
variants strengthen frozen feature transfer through synthetic
pretraining, intermediate-layer selection, token aggregation,
self-ensembling, and representation fusion~\citep{mantisv2}. UniShape
uses multiscale shape tokens and prototype-based pretraining to capture
transferable discriminative subsequences~\citep{unishape}. These
methods improve representation quality, but the target decision rule
is still learned by fitting or adapting a classification head.
Consequently, labeled support examples do not directly condition the
backbone representation of each query. ChorusTIC instead jointly
processes the labeled support set and query set and performs
classification without target-task optimization.

\paragraph{In-context classification from time series.}
Recent work replaces target-specific classifier fitting with
ICL. TIC-FM treats the target training split as
context and combines a pretrained time series encoder, a projection
adapter, and a split-masked latent-memory Transformer
~\citep{TIC-FM}. It predicts the complete query set without parameter
updates, but its encoder was developed primarily for univariate
series. TiCT is pretrained end-to-end on synthetic classification
tasks and introduces bit-based label representations and specialized
output attention to support larger class spaces~\citep{tict}. Its
synthetic task construction is based on KernelSynth and
Mixup-inspired transformations and is evaluated primarily on the
univariate UCR archive.

RocketPFN provides a concurrent route to training-free time series
classification by transforming time series into tabular features with
random convolutional kernels and applying TabPFN for in-context
classification~\citep{rocketpfn}. This two-stage formulation differs
from ChorusTIC, which integrates learned temporal and cross-channel
encoding with episodically pretrained in-context inference. iAmTime instead
adopts a broader instruction-conditioned formulation in which forecasting,
imputation, reconstruction, classification, anomaly detection, and source
separation share an encoder and decoder~\citep{iamtime}.
For classification, episode-local labels are represented as scalar
output sequences and decoded by matching the predicted value to the
nearest class code. Its pretraining mixture includes real and synthetic
sequences, including labeled series from the UCR and UEA collections,
and its classification evaluation covers selected subsets of these
archives, while its primary empirical focus is forecasting. This
setting demonstrates general instruction-conditioned task adaptation
but differs materially from ChorusTIC, which is pretrained without
real benchmark series and is designed specifically for categorical
in-context classification with learned temporal and cross-channel
interaction.

General-purpose
language models provide another training-free route. TableTime
serializes multivariate series as textual tables and combines
contextual information with neighborhood-assisted reasoning
~\citep{tabletime}. FETA retrieves exemplars independently for each
channel, asks an LLM to produce channel-level decisions, and aggregates
them through confidence-weighted late fusion~\citep{feta}. These
methods preserve training-free deployment but differ from a learned
time-series-native foundation model in representation, computational
cost, and cross-channel interaction.

\paragraph{Synthetic priors for time series models.}
Synthetic pretraining has been explored at different levels of
granularity. Chronos uses kernel-composed synthetic series to augment
large forecasting corpora~\citep{chronos}. CauKer combines Gaussian
process kernels with structural causal models to generate diverse
unlabeled sequences for representation pretraining~\citep{cauker},
while Mantis variants show that classification encoders can be
pretrained entirely on synthetic data~\citep{mantis,mantisv2}. These
approaches primarily generate individual sequences rather than
complete context and query classification tasks.

TiCT instead pretrains on synthetic binary in-context tasks constructed
by mixing two univariate KernelSynth templates, applying stochastic
time series augmentations, and assigning labels according to a
task-specific mixing threshold~\citep{tict}. TimEE constructs augmented classification tasks from the
training splits of UCR datasets~\citep{TimeEE}. A recent causal DAG
prior generates complete multivariate and multiclass datasets with
explicit temporal, cross-channel, and label structure, and validates
the prior by adapting TabPFN~\citep{o2026causal}. ChorusTIC differs by
jointly designing a classification-native temporal and cross-channel
architecture with an episodic prior aligned with its deployment
protocol. Each episode shares a task-level temporal background, while
sparse class-specific rules are applied to selected temporal regions
and channel subsets. The resulting tasks control within-channel
motifs, informative-channel selection, cross-channel phase and delay
relationships, and correlation structure, thereby directly exercising
the cross-channel evidence modeled by ChorusTIC.

\section{Detailed Method}
\label{app:method_details}

This appendix expands the method described in the main paper. It follows the
same notation and module order.

\subsection{RSSC Group Construction}
\label{app:rssc_sampling}

Consider episode $b$ with valid channel set $\mathcal{V}_b\subseteq\{1,\ldots,C_b\}$. Let $N_s=GS$ be the total number of RSSC slots. RSSC samples a flattened channel-index vector
\begin{equation}
\boldsymbol{i}_b
=
(i_{b,1},\ldots,i_{b,N_s})
\in\mathcal{V}_b^{N_s}
\end{equation}
and reshapes it into $G$ ordered groups
$\{\mathcal{I}_{b,g}\}_{g=1}^{G}$, each containing $S$ slots. The
same index tensor is shared by every context and query sample in
the episode. Consequently, each group-slot position refers to the
same source channel throughout one forward pass.

Under coverage sampling, if $|\mathcal{V}_b|\geq N_s$, we sample
without replacement:
\begin{equation}
\boldsymbol{i}_b
=
\operatorname{Perm}(\mathcal{V}_b)_{1:N_s}.
\label{eq:rssc_coverage_large_app}
\end{equation}
If $|\mathcal{V}_b|<N_s$, independent permutations are concatenated
until all slots are filled:
\begin{equation}
\begin{aligned}
R_b
&=
\left\lceil
\frac{N_s}{|\mathcal{V}_b|}
\right\rceil,\\
\boldsymbol{i}_b
&=
\left[
\operatorname{Perm}_1(\mathcal{V}_b);
\ldots;
\operatorname{Perm}_{R_b}(\mathcal{V}_b)
\right]_{1:N_s}.
\end{aligned}
\label{eq:rssc_coverage_small_app}
\end{equation}
Thus, observed channels are reused when necessary rather than
replaced by artificial zero-valued slots. The implementation also
supports independent sampling with replacement:
\begin{equation}
i_{b,n}
\overset{\mathrm{i.i.d.}}{\sim}
\operatorname{Uniform}(\mathcal{V}_b),
\qquad n=1,\ldots,N_s.
\label{eq:rssc_random_app}
\end{equation}

For sample $t$ and group $g$, the gathered raw sequence is
\begin{equation}
X_{b,t,g}
=
X_{b,t,\mathcal{I}_{b,g}}
\in\mathbb{R}^{S\times L_b}.
\label{eq:rssc_gather_app}
\end{equation}
A channel mask is used only for genuinely missing or unavailable
channels. Repeated RSSC slots remain valid observations and are not
masked.

\subsection{Patch Tokenization}
\label{app:tokenizer}

Each selected channel is linearly resampled to a common length
$L_0$. Let
$x_{b,t,g,s}\in\mathbb{R}^{L_0}$ denote the resulting sequence,
which is divided into $M$ non-overlapping patches
$x_{b,t,g,s,m}\in\mathbb{R}^{w}$ of length $w=L_0/M$.

The convolutional branches apply sequence-level normalization
\begin{equation}
\mathcal{S}(z)
=
\frac{z-\operatorname{Mean}(z)}
{\operatorname{Std}(z)+10^{-5}},
\label{eq:sequence_scaling_app}
\end{equation}
where the statistics are computed over the complete temporal axis.
The first-order difference is computed before patch aggregation:
\begin{equation}
\Delta x_{b,t,g,s}[\ell]
=
\begin{cases}
x_{b,t,g,s}[\ell+1]-x_{b,t,g,s}[\ell],
& \ell<L_0,\\
0, & \ell=L_0.
\end{cases}
\label{eq:first_difference_app}
\end{equation}

In parallel, the local mean and standard deviation are computed from
each unnormalized resampled patch:
\begin{equation}
\mu_{b,t,g,s,m}
=
\operatorname{Mean}(x_{b,t,g,s,m}),
\qquad
\sigma_{b,t,g,s,m}
=
\operatorname{Std}(x_{b,t,g,s,m}).
\label{eq:patch_stats_app}
\end{equation}

The normalized signal and its independently normalized first
difference are processed by shared convolutional encoders. Their
outputs are layer-normalized and averaged within each patch:
\begin{equation}
\begin{aligned}
h^{x}_{b,t,g,s,m}
&=
\operatorname{Pool}_{m}\!\left(
\operatorname{LN}_{x}\!\left(
\operatorname{Conv}_{x}(
\mathcal{S}(x_{b,t,g,s}))
\right)\right),\\
h^{\Delta}_{b,t,g,s,m}
&=
\operatorname{Pool}_{m}\!\left(
\operatorname{LN}_{\Delta}\!\left(
\operatorname{Conv}_{\Delta}(
\mathcal{S}(\Delta x_{b,t,g,s}))
\right)\right),
\end{aligned}
\label{eq:patch_conv_features_app}
\end{equation}
where $\operatorname{Pool}_{m}$ averages the convolutional features
assigned to patch $m$. The patch token is then
\begin{equation}
\begin{aligned}
u_{b,t,g,s,m}
=
\operatorname{Proj}\Big(
&
h^{x}_{b,t,g,s,m}
\mathbin{\Vert}
h^{\Delta}_{b,t,g,s,m}
\\
&
\mathbin{\Vert}
\operatorname{SE}_{\mu}(\mu_{b,t,g,s,m})
\mathbin{\Vert}
\operatorname{SE}_{\sigma}(\sigma_{b,t,g,s,m})
\Big),
\end{aligned}
\label{eq:patch_token_app}
\end{equation}
where $u_{b,t,g,s,m}\in\mathbb{R}^{P}$ and $\Vert$ denotes
concatenation. Stacking the tokens within sampled group $g$ gives
\begin{equation}
U_{b,t,g}^{(0)}
\in\mathbb{R}^{S\times M\times P}.
\end{equation}

\subsection{Dual-Axis Encoder and Slot Readout}
\label{app:dual_axis}

For a mini-batch of $B$ episodes, all $BTG$ sampled group instances
are processed in parallel. Let $\bar{B}=BTG$. At layer $\ell$, the
input has shape
\begin{equation}
U^{(\ell)}
\in\mathbb{R}^{\bar{B}\times S\times M\times P}.
\end{equation}

\paragraph{Temporal-axis block.}
The tensor is reshaped as
\begin{equation}
U_{\mathrm{temp}}^{(\ell)}
\in\mathbb{R}^{(\bar{B}S)\times M\times P},
\end{equation}
so each sampled channel is treated as an independent patch sequence.
The temporal block applies
\begin{equation}
\begin{aligned}
A_{\mathrm{temp}}^{(\ell)}
&=
\operatorname{MHSA}_{\mathrm{temp}}^{(\ell)}
\left(
\operatorname{LN}(U_{\mathrm{temp}}^{(\ell)})
\right),\\
\bar{U}_{\mathrm{temp}}^{(\ell)}
&=
U_{\mathrm{temp}}^{(\ell)}
+
A_{\mathrm{temp}}^{(\ell)},\\
\widetilde{U}_{\mathrm{temp}}^{(\ell)}
&=
\bar{U}_{\mathrm{temp}}^{(\ell)}
+
\operatorname{FFN}_{\mathrm{temp}}^{(\ell)}
\left(
\operatorname{LN}(\bar{U}_{\mathrm{temp}}^{(\ell)})
\right).
\end{aligned}
\label{eq:temporal_block_app}
\end{equation}

\paragraph{Channel-axis block.}
After restoring the slot and patch axes, the output is reshaped as
\begin{equation}
U_{\mathrm{chan}}^{(\ell)}
\in\mathbb{R}^{(\bar{B}M)\times S\times P}.
\end{equation}
Thus, every aligned patch position attends across the sampled slots:
\begin{equation}
\begin{aligned}
A_{\mathrm{chan}}^{(\ell)}
&=
\operatorname{MHSA}_{\mathrm{chan}}^{(\ell)}
\left(
\operatorname{LN}(U_{\mathrm{chan}}^{(\ell)});
M_{\mathrm{ch}}
\right),\\
\bar{U}_{\mathrm{chan}}^{(\ell)}
&=
U_{\mathrm{chan}}^{(\ell)}
+
A_{\mathrm{chan}}^{(\ell)},\\
U_{\mathrm{chan}}^{(\ell+1)}
&=
\bar{U}_{\mathrm{chan}}^{(\ell)}
+
\operatorname{FFN}_{\mathrm{chan}}^{(\ell)}
\left(
\operatorname{LN}(\bar{U}_{\mathrm{chan}}^{(\ell)})
\right).
\end{aligned}
\label{eq:channel_block_app}
\end{equation}
The optional mask $M_{\mathrm{ch}}$ excludes genuinely unavailable
channels from key and value positions.

After $L_D$ dual-axis layers, the refined patch tokens for slot $s$
are denoted by
$U_{b,t,g,s,:}^{(L_D)}\in\mathbb{R}^{M\times P}$. A shared
learnable summary token $q_{\mathrm{ts}}\in\mathbb{R}^{P}$ is
prepended before temporal readout:
\begin{equation}
\begin{aligned}
R_{b,t,g,s}
&=
\operatorname{Readout}
\left(
[q_{\mathrm{ts}};
U_{b,t,g,s,:}^{(L_D)}]
\right),\\
z_{b,t,g,s}
&=
R_{b,t,g,s,0}
\in\mathbb{R}^{P}.
\end{aligned}
\label{eq:slot_readout_app}
\end{equation}

\subsection{Fixed-Dimensional RSSC Composition}
\label{app:rssc_composition}

Learnable group embeddings $e_g\in\mathbb{R}^{P}$ and slot
embeddings $r_s\in\mathbb{R}^{P}$ distinguish positions in the RSSC
interface. Each slot representation is projected as
\begin{equation}
\widetilde{z}_{b,t,g,s}
=
\rho(z_{b,t,g,s}+e_g+r_s)
\in\mathbb{R}^{d_s}.
\label{eq:slot_projection_app}
\end{equation}
The final sample representation is
\begin{equation}
h_{b,t}
=
\operatorname{LN}\!\left(
\mathop{\Vert}_{g=1}^{G}
\mathop{\Vert}_{s=1}^{S}
\widetilde{z}_{b,t,g,s}
\right)
\in\mathbb{R}^{d_h},
\qquad
d_h=GSd_s.
\label{eq:rssc_output_app}
\end{equation}
Compared with global pooling, RSSC preserves the positions of
multiple sampled channel views while providing a representation width
that is independent of the original channel count.

For one episode, stacking all context and query samples gives
\begin{equation}
H
=
[H^c;H^q]
\in\mathbb{R}^{T\times d_h},
\qquad
T=N_c+N_q.
\label{eq:episode_representation_app}
\end{equation}

\subsection{Column Distribution Modeling}
\label{app:column_model}

Consider feature axis $j\in\{1,\ldots,d_h\}$. A shared scalar
projection maps each activation to a cell embedding:
\begin{equation}
q_{i,j}^{(0)}
=
f_{\mathrm{cell}}(H_{i,j})
\in\mathbb{R}^{d_e},
\qquad
Q_j^{(0)}
=
[q_{1,j}^{(0)};\ldots;q_{T,j}^{(0)}].
\label{eq:column_input_app}
\end{equation}
All parameters are shared across feature axes.

The column encoder contains $L_{\mathrm{col}}$ induced-attention
blocks. At layer $\ell$, learnable inducing tokens
$I^{(\ell)}\in\mathbb{R}^{N_{\mathrm{ind}}\times d_e}$ attend only
to the context cells:
\begin{equation}
S_j^{(\ell)}
=
\operatorname{MAB}_{1}^{(\ell)}
\left(
I^{(\ell)},
Q_{j,1:N_c}^{(\ell)},
Q_{j,1:N_c}^{(\ell)}
\right)
\in\mathbb{R}^{N_{\mathrm{ind}}\times d_e}.
\label{eq:column_bank_app}
\end{equation}
All cells then read from the context-derived bank:
\begin{equation}
Q_j^{(\ell+1)}
=
\operatorname{MAB}_{2}^{(\ell)}
\left(
Q_j^{(\ell)},
S_j^{(\ell)},
S_j^{(\ell)}
\right).
\label{eq:column_update_app}
\end{equation}

The context slice $Q_{j,1:N_c}^{(\ell)}$ depends only on context
cells at every layer. A query cell contributes only its own query
vector in Eq.~\eqref{eq:column_update_app}; it is never used as a key
or value. Therefore, queries neither modify the context-derived bank
nor communicate with one another.

The final states are decoded into cell-wise affine parameters:
\begin{equation}
\begin{aligned}
W_j
&=
\operatorname{LN}_{w}
\left(
f_w(Q_j^{(L_{\mathrm{col}})})
\right),\\
B_j
&=
\operatorname{LN}_{b}
\left(
f_b(Q_j^{(L_{\mathrm{col}})})
\right),
\end{aligned}
\label{eq:column_parameters_app}
\end{equation}
where $W_j,B_j\in\mathbb{R}^{T\times d_e}$. The calibrated embedding
of cell $(i,j)$ is
\begin{equation}
E_{i,j,:}
=
W_{i,j,:}H_{i,j}
+
B_{i,j,:}
\in\mathbb{R}^{d_e}.
\label{eq:column_affine_app}
\end{equation}
Processing every axis gives
\begin{equation}
E
\in\mathbb{R}^{T\times d_h\times d_e}.
\end{equation}
Column Distribution Modeling processes each axis independently.
Interactions among different feature axes are introduced only by the
row-wise Transformer described next.

\subsection{Row-Wise Feature Interaction}
\label{app:row_model}

For sample $i$, let
$E_i=E_{i,:,:}\in\mathbb{R}^{d_h\times d_e}$. We prepend
$K_{\mathrm{cls}}$ learned summary tokens
\begin{equation}
S_{\mathrm{cls}}
=
[s_1;\ldots;s_{K_{\mathrm{cls}}}]
\in\mathbb{R}^{K_{\mathrm{cls}}\times d_e}
\end{equation}
and apply a shared row Transformer:
\begin{equation}
R_i^{\mathrm{row}}
=
\mathcal{F}_{\mathrm{row}}
\left(
[S_{\mathrm{cls}};E_i]
\right).
\label{eq:row_transformer_app}
\end{equation}
The outputs corresponding to the summary tokens form the sample token
\begin{equation}
u_i
=
\operatorname{vec}
\left(
R_{i,1:K_{\mathrm{cls}}}^{\mathrm{row}}
\right)
\in\mathbb{R}^{d_u},
\qquad
d_u=K_{\mathrm{cls}}d_e.
\label{eq:row_output_app}
\end{equation}
Because $\mathcal{F}_{\mathrm{row}}$ is applied independently to each
row, it models interactions among feature axes without introducing
cross-sample information flow.

\subsection{Leakage-Protected In-Context Inference}
\label{app:in_context_classifier}

Label embeddings are injected only into context tokens:
\begin{equation}
\bar{u}_i
=
\begin{cases}
u_i+\mathcal{E}_y(y_i^c), & i\leq N_c,\\
u_i, & i>N_c.
\end{cases}
\label{eq:label_injection_app}
\end{equation}
Let
$\bar{U}=[\bar{u}_1;\ldots;\bar{u}_T]$. With target position $i$
and source position $j$, the additive attention mask is
\begin{equation}
M_{ij}
=
\begin{cases}
0, & j\leq N_c,\\
-\infty, & j>N_c.
\end{cases}
\label{eq:in_context_mask_app}
\end{equation}
Thus, context tokens are the only keys and values. Context tokens
attend to the context, while each query attends to the context using
its own hidden state as the attention query. The residual stream
preserves the query representation even though query tokens never
serve as keys or values.

Starting from $V^{(0)}=\bar{U}$, the in-context Transformer applies
\begin{equation}
V^{(\ell+1)}
=
\mathcal{G}_{\theta}^{(\ell)}
\left(
V^{(\ell)};M
\right),
\qquad
\ell=0,\ldots,L_{\mathrm{icl}}-1.
\label{eq:in_context_layers_app}
\end{equation}
The decoder maps the final query states to logits:
\begin{equation}
O^q
=
\operatorname{Dec}
\left(
V_{N_c+1:T}^{(L_{\mathrm{icl}})}
\right)
\in\mathbb{R}^{N_q\times K}.
\label{eq:decoder_logits_app}
\end{equation}
The class probabilities and predictions are
\begin{equation}
P^q
=
\operatorname{softmax}(O^q),
\qquad
\hat{y}_j^q
=
\arg\max_k P_{j,k}^q.
\label{eq:decoder_prediction_app}
\end{equation}

\subsection{Complete Synthetic Episodic Prior}
\label{app:synthetic_prior}

\paragraph{Episode configuration.}
Each episode samples
\begin{equation}
\omega
=
(K,C,L,N_c,N_q,\kappa,r,d),
\qquad
\kappa\in\{\mathrm{uni},\mathrm{multi}\},
\label{eq:prior_episode_configuration}
\end{equation}
where $\kappa$ denotes the task type, $r$ denotes the
discriminative rule family, and $d$ controls task difficulty.
For the reported model, univariate and multivariate episodes are
sampled with probabilities $0.2$ and $0.8$, respectively.
Univariate episodes use $C=1$ and
$r\in\mathcal R_{\mathrm{temp}}$, whereas multivariate episodes
use $2\leq C\leq 10$ and
$r\in\mathcal R_{\mathrm{cross}}$. Class proportions are sampled as
\[
\boldsymbol{\varrho}
\sim
\operatorname{Dirichlet}(\alpha\mathbf{1}_K),
\]
subject to every active class being represented in the context set.

\paragraph{Shared temporal background.}
A generator family is first selected from a categorical mixture:
\begin{equation}
a
\sim
\operatorname{Categorical}(\boldsymbol{\lambda}),
\qquad
W
\sim
\mathcal{P}_{a},
\qquad
W\in\mathbb{R}^{C\times L}.
\label{eq:background_mixture_app}
\end{equation}
The collection $\{\mathcal{P}_{a}\}$ contains:

\begin{itemize}
    \item smooth, periodic, and colored-noise processes;
    \item structural channel graphs with lagged or nonlinear
    dependencies;
    \item regime-switching and changepoint processes;
    \item event, spike, burst, and plateau processes;
    \item amplitude- and frequency-modulated sinusoids; and
    \item audio-like multiscale processes.
\end{itemize}

The sampled background is normalized channel-wise using robust
location and scale statistics and clipped for numerical stability.
All classes within an episode share the same background, so class
identity cannot be inferred from independently generated nuisance
dynamics.

\paragraph{Class-specific discriminative rules.}
For class $k$, we sample an informative channel subset
$\mathcal{S}_k$, an informative temporal region $\mathcal{T}_k$, and
rule parameters $\eta_k$. The class prototype is
\begin{equation}
P_k
=
\Gamma_{r,k}
\left(
W;\mathcal{S}_k,\mathcal{T}_k,\eta_k
\right),
\qquad
k=1,\ldots,K.
\label{eq:rule_operator_app}
\end{equation}
The modification is sparse in time and, for multivariate episodes,
sparse in channels.

For univariate episodes, the rule families include motif shape,
polarity, position, order, and localized deviations. Multivariate
episodes define class differences through informative-channel
selection, channel-specific motifs, relative delays, phase
relationships, and correlation regimes.

\paragraph{Instance-level variation.}
An observed instance of class $y_i$ is generated as
\begin{equation}
X_i
=
\mathcal{A}_{\xi_i}(P_{y_i})
+
\varepsilon_i,
\label{eq:nuisance_transformation_app}
\end{equation}
where $\mathcal{A}_{\xi_i}$ is an instance-specific nuisance
transformation and $\varepsilon_i$ denotes sensor noise. The
transformation family contains temporal shifts, elastic warping,
local masking, length perturbations, burst noise, quantization,
amplitude clipping, and distractor-channel perturbations. Easy
episodes use stronger discriminative rules and weaker nuisance
transformations, whereas hard episodes reduce the class margin and
increase nuisance severity.

\paragraph{Context and query construction.}
After instance generation, samples are independently shuffled and
divided into context and query sets. An episode-specific random
bijection
\begin{equation}
\sigma_{\mathcal{E}}
:
\{1,\ldots,K\}
\rightarrow
\{1,\ldots,K\}
\end{equation}
is applied to every context and query label:
\begin{equation}
\widetilde{y}
=
\sigma_{\mathcal{E}}(y).
\label{eq:episode_label_permutation_app}
\end{equation}
The model observes
$(X^c,\widetilde{Y}^c,X^q)$ but not $\widetilde{Y}^q$.

Let
\begin{equation}
\widetilde{\mathcal{C}}_{\mathcal{E}}
=
\left\{
(X_i^c,\sigma_{\mathcal{E}}(y_i^c))
\right\}_{i=1}^{N_c}.
\end{equation}
The pretraining objective is
\begin{equation}
\mathcal{L}(\Theta)
=
-
\mathbb{E}_{\mathcal{E}\sim p_{\mathrm{syn}}}
\left[
\frac{1}{N_q}
\sum_{j=1}^{N_q}
\log
p_{\Theta}
\left(
\sigma_{\mathcal{E}}(y_j^q)
\mid
X_j^q,
\widetilde{\mathcal{C}}_{\mathcal{E}}
\right)
\right].
\label{eq:training_objective_app}
\end{equation}
The episode-specific permutation prevents fixed numerical labels from
becoming associated with particular synthetic rules.

\subsection{Deployment-Time Ensembling}
\label{app:inference_ensemble}

\paragraph{Label permutation ensemble.}
For a target task with $K$ classes, define the $m$-th cyclic
permutation as
\begin{equation}
\pi_m(y)
=
1+\big((y-1+m)\bmod K\big),
\qquad
m=0,\ldots,M_{\pi}-1.
\label{eq:cyclic_permutation_app}
\end{equation}
Let $P_m\in\{0,1\}^{K\times K}$ denote its permutation matrix:
\begin{equation}
(P_m)_{y,\pi_m(y)}=1.
\label{eq:permutation_matrix_app}
\end{equation}
The model predicts using the permuted context labels:
\begin{equation}
O^{q,(m)}
=
ChorusTIC
\left(
X^c,\pi_m(Y^c),X^q
\right).
\label{eq:permuted_prediction_app}
\end{equation}
Because column $\pi_m(y)$ corresponds to original class $y$, the
logits are restored by
\begin{equation}
\widetilde{O}^{q,(m)}
=
O^{q,(m)}P_m^{\top}.
\label{eq:aligned_logits_app}
\end{equation}
The ensemble logits are
\begin{equation}
\bar{O}^{q}
=
\frac{1}{M_{\pi}}
\sum_{m=0}^{M_{\pi}-1}
\widetilde{O}^{q,(m)}.
\label{eq:label_ensemble_app}
\end{equation}

\paragraph{RSSC sampling ensemble.}
Because RSSC samples channel groups stochastically, deployment averages
predictions over $M_R$ independent RSSC draws. Let
$O^{q,(m,a)}$ denote the logits obtained using label permutation $m$
and RSSC draw $a$, where $a=1,\ldots,M_R$. The combined estimator is
\begin{equation}
\bar{O}^{q}
=
\frac{1}{M_{\pi}M_R}
\sum_{m=0}^{M_{\pi}-1}
\sum_{a=1}^{M_R}
O^{q,(m,a)}P_m^{\top}.
\label{eq:combined_ensemble_app}
\end{equation}
Within each draw, the same sampled RSSC channel indices are shared
across all context and query samples.

The final probabilities and predictions are
\begin{equation}
P^q
=
\operatorname{softmax}\left(\frac{\bar{O}^q}{\tau}\right),
\qquad
\widehat{y}_j^q
=
\arg\max_k P_{j,k}^q,
\label{eq:ensemble_prediction_app}
\end{equation}
where the temperature is set to $\tau=0.9$ by default. The reported
configuration uses $M_\pi=8$ cyclic label permutations and $M_R=4$
independent RSSC draws, resulting in 32 ensemble members per
prediction.

\subsection{Hierarchical Extension for Many-Class Tasks}
\label{app:many_class}

The native decoder supports at most $K_{\max}$ classes. For $K>K_{\max}$, we construct a balanced tree whose leaves correspond to the original classes and whose internal nodes have at most $K_{\max}$ children.

For internal node $v$, let $\operatorname{ch}(v)$ denote its child groups and let $\mathcal{C}_v$ contain the context samples whose labels belong to descendants of $v$. The original labels in
$\mathcal{C}_v$ are replaced by local child-group indices. The model then predicts
\begin{equation}
p_{\Theta}
\left(
g
\mid
X,\mathcal{C}_v
\right),
\qquad
g\in\operatorname{ch}(v).
\label{eq:node_routing_app}
\end{equation}
At the final internal node, each child corresponds to an individual
class. For class $y$, let $v_0,\ldots,v_{D_y-1}$ denote the internal nodes on its path and $g_{v_d}(y)$ the child selected at node $v_d$. Its probability is
\begin{equation}
p_{\Theta}
\left(
y\mid X,\mathcal{C}_{\tau}
\right)
=
\prod_{d=0}^{D_y-1}
p_{\Theta}
\left(
g_{v_d}(y)
\mid
X,\mathcal{C}_{v_d}
\right).
\label{eq:hierarchical_probability_app}
\end{equation}
This procedure decomposes a many-class task into a sequence of
native-capacity in-context decisions and requires no target-task
parameter updates.

\section{Reproducibility Details}
\label{sec:reproducibility}

\subsection{Datasets and Evaluation Splits}
\label{subsec:repro_datasets}

We evaluate on all 128 datasets in the UCR Time Series
Classification Archive and all 30 datasets in the UEA Multivariate
Time Series Classification Archive. We use the official train/test
splits without excluding datasets or modifying their labels.  Both archives are publicly available from their official repositories.

For full-context evaluation, the complete official training split is
provided to ChorusTIC as labeled context, and the complete official
test split is used as the query set. No ChorusTIC parameter is updated
on a target dataset. For frozen-representation baselines, the
pretrained backbone remains fixed, while the specified lightweight
classifier is fitted using only the official target training split.
Test labels are used only for final accuracy computation.

For fixed-shot evaluation, we sample $k\in\{5,10\}$ labeled
examples per class from the official training split and use the
complete test split as the query set. A dataset is excluded from the
$k$-shot setting only if at least one class contains fewer than $k$
training examples. Results are averaged over five independently
sampled support sets, and all methods use identical support sets for
each dataset and label budget. The sampling seeds are $0$, $1$, $2$,
$3$, and $4$. For proportional-label evaluation, we retain 20\%,
30\%, 40\%, 50\%, or 60\% of the official training split using
class-stratified sampling, with identical sampled subsets shared
across methods.

No real-world time series are used to pretrain ChorusTIC. Synthetic episodes are generated online from the episodic prior
described in Section B.8 and contain no samples
from the UCR or UEA archives.

\subsection{Input Preprocessing}
\label{subsec:repro_preprocessing}

All preprocessing is performed independently for each dataset. UCR
samples are treated as univariate time series and represented with a
singleton channel axis, whereas UEA samples retain their original
multivariate organization. We do not flatten or concatenate UEA
channels before ChorusTIC encoding. No dataset-level channel selection
or truncation is applied during preprocessing; RSSC subsequently
samples channel groups within the ChorusTIC encoder.

The data reader maps every input to the model length
$L_0=512$. Time series with a different original length are resampled
along the temporal axis using linear interpolation with
\texttt{align\_corners=False}. Missing values in UEA files are
replaced with zero before temporal interpolation. The same
preprocessing procedure is applied to the official training and test
splits, without using test labels or test-set statistics. Labels are
mapped to consecutive integers using a label encoder fitted on the
training split and reused for the corresponding test split.

\subsection{ChorusTIC Configuration}
\label{subsec:repro_chorustic_config}

The reported model uses the checkpoint at pretraining step 6000. Its
architecture is reconstructed from
\texttt{model\_hparams\_latest.json}, and checkpoint loading is
performed with strict consistency checks for both the RSSC encoder
and the in-context learner. All parameters are set to evaluation mode
and remain frozen throughout target-task evaluation.

Table~\ref{tab:chorustic_architecture_config} lists the final model configuration. The final inference configuration is given in
Table~\ref{tab:chorustic_inference_config}. Batch-size parameters
control memory consumption only. When a CUDA out-of-memory error is
detected, the implementation reduces the relevant batch sizes and
retries the same computation. The effective batch sizes are recorded
in the output files.

\begin{table}[!htbp]
  \centering
  \caption{\textbf{Final ChorusTIC architecture and pretraining
  configuration.} All values correspond to the checkpoint used for
  the reported UCR and UEA results.}
  \label{tab:chorustic_architecture_config}
  \footnotesize
  \setlength{\tabcolsep}{3.5pt}
  \renewcommand{\arraystretch}{1.02}
  \begin{tabularx}{\textwidth}{@{}Xl@{\hspace{1.4em}}Xl@{}}
    \toprule
    Configuration & Value & Configuration & Value \\
    \midrule
    Input length $L_0$                    & 512
      & Task-level embedding width          & 128 \\
    Number of temporal patches $M$        & 32
      & Column-attention blocks             & 3 \\
    RSSC groups $G$                       & 4
      & Column-attention heads              & 4 \\
    Slots per RSSC group $S$              & 4
      & Column inducing tokens              & 128 \\
    RSSC slot dimension                   & 32
      & Row-interaction blocks              & 3 \\
    RSSC sampling strategy                & Coverage
      & Row-attention heads                 & 8 \\
    Signal-encoder width                  & 512
      & Row summary tokens                 & 4 \\
    Dual-axis encoder layers              & 3
      & ICL Transformer blocks             & 12 \\
    Temporal-attention heads              & 8
      & ICL attention heads                & 4 \\
    Channel-attention heads               & 4
      & ICL feed-forward expansion         & 2 \\
    Temporal feed-forward width           & 512
      & ICL dropout                        & 0 \\
    Channel feed-forward width            & 512
      & Maximum native class count $K_{\max}$ & 10 \\
    Dual-axis dropout                     & 0.1
      & Pretraining optimizer              & AdamW \\
    \midrule
    Learning rate                         & $1\times10^{-4}$
      & Weight decay                        & 0 \\
    Episode batch size                    & 36
      & Pretraining steps                   & 6000 \\
    Gradient clipping                     & 1.0
      & Numerical precision                & FP32 with AMP \\
    \bottomrule
  \end{tabularx}
\end{table}

\begin{table}[!htbp]
  \centering
  \caption{\textbf{Final ChorusTIC inference configuration.}}
  \label{tab:chorustic_inference_config}
  \small
  \setlength{\tabcolsep}{5pt}
  \renewcommand{\arraystretch}{1.05}
  \begin{tabular}{@{}ll@{}}
    \toprule
    Parameter & Final value \\
    \midrule
    Checkpoint                             & step-6000 \\
    Evaluation mode                        & classifier\_v2 \\
    Context mode                           & full training split \\
    Label-permutation ensemble size                       & 8 \\
    Cyclic  label-permutation ensemble                   & enabled \\
    RSSC inference draws $M_R$             & 4 \\
    Softmax temperature                    & 0.9 \\
    Channel selection                      & disabled \\
    Evaluation seed             & 0 \\
    Total ensemble evaluations  & 32 \\
    Hierarchical classification & Enabled for $K>10$ \\
    \bottomrule
  \end{tabular}
\end{table}

\subsection{Hyperparameter Development and Baseline Configuration}
\label{subsec:repro_hyperparameters}

We distinguish prediction-relevant hyperparameters from parameters
that affect only computational batching. The latter, including
\texttt{v2\_batch\_size}, \texttt{mantis\_batch\_size}, and the
ensemble batch sizes, are adjusted according to available GPU memory
and do not change the prediction rule.

During preliminary development, prediction-relevant settings are
evaluated on a fixed synthetic validation set containing 64 episodes
sampled independently from the episodic prior. UCR and UEA test
labels are not used for hyperparameter selection. Ensemble sizes are
selected by considering validation accuracy and inference cost, with
larger settings omitted once accuracy gains begin to saturate. Table~\ref{tab:hyperparameter_development}
summarizes the candidate values and final settings.

\begin{table}[!htbp]
  \centering
  \caption{\textbf{Development ranges and final inference
  hyperparameters.}}
  \label{tab:hyperparameter_development}
  \scriptsize
  \setlength{\tabcolsep}{3pt}
  \renewcommand{\arraystretch}{1.08}
  \begin{tabular}{@{}p{0.23\linewidth}p{0.25\linewidth}
                  p{0.12\linewidth}p{0.30\linewidth}@{}}
    \toprule
    Parameter & Values considered & Final & Selection criterion \\
    \midrule
    Label-permutation ensemble size $M_\pi$
      & $\{1,2,4,8\}$
      & 8
      & Validation accuracy and inference cost \\
    RSSC inference draws $M_R$
      & $\{1,2,4,8\}$
      & 4
      & Validation accuracy and inference cost \\
    Cyclic label permutation
      & Enabled, disabled
      & Enabled
      & Validation accuracy \\
    \bottomrule
  \end{tabular}
\end{table}

For all comparison methods, we use the authors' released
implementations, pretrained checkpoints, preprocessing procedures,
and recommended default hyperparameters. Frozen foundation models
use the representation readout and target-classifier protocol
specified in their original implementations or papers. No baseline
is tuned separately on a target test split.

\begin{table}[!htbp]
  \centering
  \caption{\textbf{Implementations and target-task protocols of the
  comparison methods.} ``Official'' indicates the use of an
  author-released implementation.}
  \label{tab:baseline_reproducibility}
  \scriptsize
  \setlength{\tabcolsep}{3pt}
  \renewcommand{\arraystretch}{1.08}
  \begin{tabular}{@{}p{0.18\linewidth}p{0.13\linewidth}
                  p{0.39\linewidth}p{0.20\linewidth}@{}}
    \toprule
    Method & Official & Version or checkpoint & Target protocol \\
    \midrule
    MOMENT
      & Yes
      & \texttt{MOMENT-1-base}
      & Frozen feature + SVM \\
    Mantis
      & Yes
      & \texttt{Mantis-8M}
      & Frozen feature + RF \\
    MantisV2
      & Yes
      & \texttt{MantisV2}
      & Frozen feature + LR/RF \\
    UniShape
      & Yes
      & \texttt{unishape\_checkpoint\_zeroshot}
      & Frozen feature + RF \\
    NuTime
      & Yes
      & \texttt{checkpoint\_bias9}
      & Frozen CLS + RF \\
    TabICL
      & Yes
      & \texttt{tabicl-classifier-v1.1-20250506}
      & In-context inference \\
    TabICLv2
      & Yes
      & \texttt{tabicl-classifier-v2-20260212}
      & In-context inference \\
    TIC-FM
      & Yes
      &  \texttt{TIC-FM}
      & In-context inference \\
    TiCT
      & Yes
      & ResNet, 47M parameters
      & In-context inference \\
    \bottomrule
  \end{tabular}
\end{table}

\FloatBarrier

\subsection{Evaluation Metrics and Statistical Analysis}
\label{subsec:repro_metrics}

For dataset $d$, classification accuracy is
\begin{equation}
  \operatorname{Acc}_d
  =
  \frac{1}{N_d}
  \sum_{i=1}^{N_d}
  \mathbb{I}\!\left[
    \widehat{y}_{d,i}=y_{d,i}
  \right],
  \label{eq:repro_accuracy}
\end{equation}
where $N_d$ is the number of test samples. For an archive containing
$D$ datasets, average accuracy is the unweighted macro-average
\begin{equation}
  \operatorname{AvgAcc}
  =
  \frac{1}{D}
  \sum_{d=1}^{D}
  \operatorname{Acc}_d.
  \label{eq:repro_avg_accuracy}
\end{equation}
Macro-averaging gives equal weight to every dataset and prevents
large datasets from dominating the archive-level result.

For average rank, methods are ranked separately within each dataset,
with rank 1 assigned to the highest accuracy. Tied methods receive
their average rank. The reported average rank is the arithmetic mean
of these per-dataset ranks. Rankings, best/second-best markings, and
win/tie/loss counts are computed from the stored full-precision
accuracies rather than the rounded values displayed in the tables.

Win/tie/loss counts are reported from the perspective of ChorusTIC.
A win indicates
\(
\operatorname{Acc}_{d,\mathrm{ChorusTIC}}
>
\operatorname{Acc}_{d,\mathrm{baseline}}
\),
a loss indicates the opposite, and exact equality is counted as a
tie.

Relative improvement over a reference method is computed from the
unrounded macro-average accuracies as
\begin{equation}
  \operatorname{RelGain}
  =
  \frac{
    \operatorname{AvgAcc}_{\mathrm{ours}}
    -
    \operatorname{AvgAcc}_{\mathrm{ref}}
  }{
    \operatorname{AvgAcc}_{\mathrm{ref}}
  }
  \times 100\%.
  \label{eq:repro_relative_gain}
\end{equation}

\paragraph{Paired statistical testing.}
To assess whether the observed performance differences are
statistically reliable across datasets, we compare ChorusTIC with
each baseline using a two-sided Wilcoxon signed-rank test on paired
per-dataset accuracies. UEA-30 and UCR-128 are analyzed separately
because they represent distinct benchmark collections and evaluation
settings. For each comparison, the null hypothesis is that the
distribution of the nonzero paired accuracy differences is symmetric
about zero.

Datasets with an exact zero difference are omitted from the
signed-rank calculation, following the standard Wilcoxon zero-difference
convention. The resulting number of nonzero paired differences is
reported as $n_{\mathrm{eff}}$. Let $W^{+}$ and $W^{-}$ denote the
sums of the ranks associated with positive and negative differences,
respectively. The reported two-sided test statistic is
\begin{equation}
  W
  =
  \min\!\left(W^{+},W^{-}\right).
  \label{eq:wilcoxon_statistic}
\end{equation}

Because ChorusTIC is compared with multiple baselines, the resulting
$p$-values are adjusted using the Holm procedure. Correction is
performed separately within each archive: the UEA-30 family contains
eight baseline comparisons, whereas the UCR-128 family contains ten.
Statistical significance is assessed at $\alpha=0.05$ using the
Holm-corrected $p$-values. All tests use stored full-precision
per-dataset accuracies, and statistical conclusions are based on the
corrected rather than the uncorrected values.

\subsection{Computing Infrastructure}
\label{subsec:repro_compute}

Table~\ref{tab:computing_environment} reports the computing
and software environment used for pretraining and evaluation.

\begin{table}[!htbp]
  \centering
  \caption{\textbf{Computing and software environment.}}
  \label{tab:computing_environment}
  \small
  \setlength{\tabcolsep}{5pt}
  \renewcommand{\arraystretch}{1.05}
  \begin{tabular}{@{}ll@{}}
    \toprule
    Item & Configuration \\
    \midrule
    GPU model and count
      & $4\times$ NVIDIA Tesla V100 PCIe \\
    GPU memory
      & 32\,GiB per GPU (128\,GiB total) \\
    GPU driver
      & 570.86.15 \\
    CPU model
      & Intel Xeon Gold 6140 @ 2.30\,GHz \\
    System memory
      & 251\,GiB \\
    Operating system
      & Ubuntu 18.04.6 LTS \\
    Python
      & 3.10.18 \\
    PyTorch
      & 2.5.1 \\
    CUDA runtime
      & 12.4 \\
    cuDNN
      & 9.1.0 \\
    NumPy
      & 2.0.1 \\
    scikit-learn
      & 1.7.2 \\
    \bottomrule
  \end{tabular}
\end{table}
\FloatBarrier



\section{Additional Experiments}
\label{sec:APPexperiments}

\subsection{Evaluation Details}
\label{subsec:exp_setup}

\paragraph{Benchmarks.}
We use the official train/test splits of the UEA-30 multivariate
archive~\citep{uea} and the UCR-128 univariate
archive~\citep{dau2019ucr}. For ChorusTIC, the training split supplies labeled
context and the test split forms the query set; no target-task parameter is
updated. In fixed-shot evaluation, we sample 5 or 10 context examples per
class and retain the complete test split. Results are averaged over five
independently sampled context sets shared across methods. A dataset is included
at a given shot level only when every class contains enough training examples.

\paragraph{Baselines.}
On UCR, TIC-FM~\citep{TIC-FM} and TiCT~\citep{tict} provide the closest
time-series ICL comparisons. TabICL~\citep{tabicl} and
TabICLv2~\citep{tabiclv2} serve as generic ICL controls after each series is
vectorized. Frozen TSFM baselines include MOMENT~\citep{moment},
Mantis~\citep{mantis}, MantisV2~\citep{mantisv2},
UniShape~\citep{unishape}, and NuTime~\citep{nutime}. Each TSFM backbone remains
fixed, and its default final-layer readout is used to fit the lightweight
classifier named in the tables. NuTime+RF uses the final normalized CLS
representation and a random forest.

\paragraph{Metrics.}
We report the unweighted average accuracy and average rank across datasets.
Win/tie/loss counts use paired per-dataset accuracies and are reported from the
perspective of ChorusTIC. The five-run average applies to sampled
low-label context sets.

\subsection{Archive-Level Results}
\label{app:main_results}

\paragraph{UEA-30}
\label{app:uea_results}

Table~\ref{tab:uea_main_1} supplements the main-paper comparison with
win/tie/loss counts. ChorusTIC is the only evaluated method that combines
training-free deployment with native multivariate encoding. Its gains over
generic ICL controls are consistent with the need to preserve temporal and
cross-channel structure rather than treating each series as an unordered
feature vector.

\begin{table}[!htbp]
  \centering
  \caption{
  \textbf{Classification results on the complete UEA-30 archive.}
  ``Target fit'' indicates whether a dataset-specific classifier is fitted on
  the target training split. W/T/L counts are reported from the perspective of
  ChorusTIC. Best and second-best average accuracies and average ranks are
  shown in \textbf{bold} and \underline{underlined}, respectively.
  }
  \label{tab:uea_main_1}
  \scriptsize
  \setlength{\tabcolsep}{4.2pt}
  \renewcommand{\arraystretch}{1.06}
  \begin{tabular*}{0.82\linewidth}
    {@{\extracolsep{\fill}}llcrrr@{}}
    \toprule
    Protocol
      & Method
      & \shortstack{Target\\fit}
      & \shortstack{Avg.\\Acc. $\uparrow$}
      & \shortstack{Avg.\\Rank $\downarrow$}
      & W/T/L \\
    \midrule

    Time-series ICL
      & ChorusTIC
      & No
      & \textbf{72.27\%}
      & \textbf{3.57}
      & \textemdash \\

    \midrule

    \multirow{2}{*}{Generic ICL}
      & TabICL
      & No
      & 65.33\%
      & 5.18
      & 18/3/9 \\

      & TabICLv2
      & No
      & 67.96\%
      & 4.37
      & 16/1/13 \\

    \midrule

    \multirow{6}{*}{Frozen TSFM}
      & MOMENT+SVM
      & Yes
      & 68.17\%
      & 5.48
      & 22/2/6 \\

      & Mantis+RF
      & Yes
      & 69.34\%
      & 5.22
      & 21/1/8 \\

      & MantisV2+LR
      & Yes
      & \underline{70.50\%}
      & \underline{4.02}
      & 18/3/9 \\

      & MantisV2+RF
      & Yes
      & 69.54\%
      & 4.63
      & 16/2/12 \\

      & UniShape+RF
      & Yes
      & 69.72\%
      & 4.92
      & 18/3/9 \\

      & NuTime+RF
      & Yes
      & 57.92\%
      & 7.62
      & 26/1/3 \\

    \bottomrule
  \end{tabular*}
\end{table}

\begin{table}[!htbp]
  \centering
  \caption{
  \textbf{Classification results on the complete UCR-128 archive.}
  ``Target fit'' indicates whether a classifier is fitted on the target
  training split. Best and second-best results are shown in \textbf{bold} and
  \underline{underlined}, respectively.
  }
  \label{tab:ucr_main}
  \scriptsize
  \setlength{\tabcolsep}{4.5pt}
  \renewcommand{\arraystretch}{1.04}
  \begin{tabular*}{0.82\linewidth}
    {@{\extracolsep{\fill}}llcrr@{}}
    \toprule
    Protocol & Method & Target fit & Avg. Acc. & Avg. Rank \\
    \midrule
    \multirow{6}{*}{Frozen TSFM}
      & MOMENT+SVM     & Yes & 77.98\% & 6.11 \\
      & Mantis+RF      & Yes & 78.67\% & 6.42 \\
      & MantisV2+RF    & Yes & 78.79\% & 6.51 \\
      & MantisV2+LR    & Yes & \underline{80.03\%} & 5.50 \\
      & UniShape+RF    & Yes & 78.86\% & 5.83 \\
      & NuTime+RF      & Yes & 69.39\% & 9.55 \\
    \midrule
    \multirow{2}{*}{Generic ICL}
      & TabICL         & No & 76.83\% & 6.38 \\
      & TabICLv2       & No & 78.88\% & 5.15 \\
    \midrule
    \multirow{3}{*}{Time-series ICL}
      & TiCT           & No & 79.17\% & \underline{4.81} \\
      & TIC-FM         & No & 80.01\% & 5.32 \\
      & ChorusTIC      & No & \textbf{81.16\%} & \textbf{4.43} \\
    \bottomrule
  \end{tabular*}
\end{table}

\paragraph{UCR-128}
\label{app:ucr_results}

Table~\ref{tab:ucr_main} summarizes the complete UCR-128 comparison.
The same pretrained ChorusTIC model transfers to the single-channel
setting and achieves the best aggregate accuracy and rank without fitting
a target-specific classifier.

\subsection{Per-Dataset Results}
\label{subsec:per_dataset_results}

\paragraph{UEA-30}

Table~\ref{tab:uea_full_results_9methods} reports the per-dataset
accuracies underlying the UEA aggregate statistics. It exposes variation
across heterogeneous multivariate tasks while confirming ChorusTIC's best
archive-level average accuracy and rank without target-task fitting.

\paragraph{UCR-128}

Table~\ref{tab:ucr_full_results_11methods} provides the corresponding
per-dataset results across all 128 tasks. These results show the variation
behind the archive averages while retaining the same training-free
classification protocol for ChorusTIC.

\paragraph{Statistical Comparison Results}

We conduct two-sided Wilcoxon signed-rank tests using full-precision
per-dataset accuracies, excluding zero paired differences from the
signed-rank calculation. Holm correction is applied separately to
the UEA-30 and UCR-128 families of comparisons. On UEA-30, the
differences remain significant after correction for MOMENT+SVM
(W/T/L $=22/2/6$, raw $p=4.73\times10^{-4}$, adjusted
$p=3.31\times10^{-3}$) and NuTime+RF
(W/T/L $=26/1/3$, raw $p=4.70\times10^{-6}$, adjusted
$p=3.76\times10^{-5}$). On UCR-128, the differences remain
significant after correction for MOMENT+SVM, Mantis+RF,
MantisV2+RF, UniShape+RF, NuTime+RF, and TabICL. For example,
Mantis+RF has a W/T/L count of $93/4/31$ with adjusted
$p=5.22\times10^{-6}$, NuTime+RF has $121/1/6$ with adjusted
$p=4.90\times10^{-30}$, and TabICL has $76/6/46$ with adjusted
$p=2.93\times10^{-3}$.


\begin{table}[!htbp]
  \centering
  \caption{
  \textbf{Per-dataset classification accuracy on the complete UEA-30 archive.}
  Best and second-best results within each dataset are shown in \textbf{bold}
  and \underline{underlined}, respectively, based on the displayed four-decimal
  accuracies. Avg. Acc. is the macro-average over datasets; Avg. Rank is
  computed among the 9 displayed methods using full-precision accuracies.
  Abbreviations: MOMENT=MOMENT+SVM, Mantis=Mantis+RF, MV2-LR=MantisV2+LR,
  MV2-RF=MantisV2+RF, UniShape=UniShape+RF, and NuTime=NuTime+RF.
  }
  \label{tab:uea_full_results_9methods}
  \scriptsize
  \setlength{\tabcolsep}{2.8pt}
  \renewcommand{\arraystretch}{1.03}
  \resizebox{\textwidth}{!}{%
  \begin{tabular}{@{}l*{9}{r}@{}}
    \toprule
    Dataset & TabICL & TabICLv2 & MOMENT & Mantis & MV2-LR & MV2-RF & UniShape & NuTime & ChorusTIC \\
    \midrule
    ArticularyWordRecognition & 0.9800 & 0.9833 & 0.9600 & \underline{0.9927} & \textbf{0.9933} & \textbf{0.9933} & 0.9900 & 0.7600 & 0.9800 \\
    AtrialFibrillation & 0.2000 & \textbf{0.3333} & 0.1333 & \underline{0.2800} & 0.1333 & 0.0667 & 0.2000 & 0.1333 & 0.2667 \\
    BasicMotions & \textbf{1.0000} & \underline{0.9750} & \textbf{1.0000} & \textbf{1.0000} & \textbf{1.0000} & \textbf{1.0000} & \textbf{1.0000} & \textbf{1.0000} & \textbf{1.0000} \\
    CharacterTrajectories & 0.9847 & \textbf{0.9937} & 0.9742 & 0.9401 & 0.9742 & 0.9568 & 0.9749 & 0.8621 & \underline{0.9889} \\
    Cricket & 0.9444 & 0.9444 & \underline{0.9861} & \textbf{1.0000} & \underline{0.9861} & \underline{0.9861} & 0.9722 & 0.8611 & 0.9722 \\
    DuckDuckGeese & 0.2000 & 0.2000 & 0.4600 & 0.3880 & \underline{0.4800} & \textbf{0.5000} & 0.4600 & 0.2600 & 0.4600 \\
    ERing & \underline{0.9630} & 0.9556 & 0.9185 & 0.9452 & \textbf{0.9704} & 0.9519 & 0.9481 & 0.7185 & 0.9185 \\
    EigenWorms & 0.4198 & 0.4198 & 0.7863 & 0.7252 & \underline{0.8397} & 0.7863 & 0.7786 & 0.5191 & \textbf{0.9008} \\
    Epilepsy & 0.9203 & 0.9493 & 0.9855 & \underline{0.9957} & \textbf{1.0000} & \textbf{1.0000} & \textbf{1.0000} & 0.9710 & \textbf{1.0000} \\
    EthanolConcentration & \underline{0.4297} & \textbf{0.4449} & 0.2700 & 0.2837 & 0.3764 & 0.3916 & 0.4030 & 0.3536 & 0.3802 \\
    FaceDetection & \underline{0.6107} & \textbf{0.6544} & 0.5360 & 0.5178 & 0.5292 & 0.5184 & 0.5647 & 0.5258 & 0.5573 \\
    FingerMovements & 0.5600 & 0.5500 & \textbf{0.6000} & 0.5160 & \underline{0.5700} & 0.5300 & 0.5500 & 0.5000 & 0.4600 \\
    HandMovementDirection & \textbf{0.4595} & \underline{0.4459} & 0.3378 & 0.2730 & 0.2432 & 0.2432 & 0.3108 & 0.2432 & 0.2838 \\
    Handwriting & 0.2635 & 0.2824 & 0.3106 & 0.3369 & \textbf{0.3635} & 0.2906 & 0.2753 & 0.1400 & \underline{0.3612} \\
    Heartbeat & 0.7317 & 0.7610 & 0.7122 & 0.7678 & 0.7854 & \textbf{0.8195} & 0.7854 & 0.6098 & \underline{0.8000} \\
    InsectWingbeatSubset & 0.4370 & 0.5400 & 0.4490 & 0.5396 & \textbf{0.6210} & 0.5410 & 0.3730 & 0.1940 & \underline{0.6140} \\
    JapaneseVowels & \textbf{0.9784} & 0.9730 & 0.7811 & 0.9697 & 0.9730 & 0.9514 & 0.9270 & 0.8541 & \underline{0.9757} \\
    LSST & 0.5799 & \underline{0.6833} & 0.6557 & 0.6038 & 0.6221 & 0.6330 & 0.6326 & 0.3564 & \textbf{0.6849} \\
    Libras & 0.8944 & 0.8944 & 0.8778 & 0.8911 & \textbf{0.9222} & 0.8778 & 0.8611 & 0.7833 & \underline{0.9111} \\
    MotorImagery & 0.5000 & 0.5000 & 0.4700 & 0.5220 & 0.5000 & 0.5600 & \underline{0.5700} & \textbf{0.6400} & 0.5000 \\
    NATOPS & 0.8667 & \textbf{0.9556} & 0.8167 & 0.8556 & 0.8611 & \underline{0.8778} & 0.8500 & 0.6667 & 0.8722 \\
    PEMS-SF & 0.1734 & 0.1734 & 0.7630 & \underline{0.9098} & 0.8266 & \textbf{0.9884} & \textbf{0.9884} & 0.3699 & 0.8613 \\
    PenDigits & 0.9831 & \underline{0.9834} & \textbf{0.9837} & 0.9410 & 0.9688 & 0.9431 & 0.9548 & 0.8974 & 0.9757 \\
    PhonemeSpectra & 0.1566 & 0.1163 & 0.2648 & 0.2746 & \underline{0.2997} & 0.2815 & 0.2434 & 0.1452 & \textbf{0.3185} \\
    RacketSports & 0.8553 & 0.8553 & 0.8158 & \textbf{0.9276} & \underline{0.9013} & 0.8882 & 0.8618 & 0.7368 & 0.8684 \\
    SelfRegulationSCP1 & \underline{0.8805} & \textbf{0.9044} & 0.7713 & 0.8061 & 0.7782 & 0.7782 & 0.7952 & 0.7747 & 0.8191 \\
    SelfRegulationSCP2 & 0.4944 & 0.4889 & \underline{0.5389} & 0.4922 & 0.5111 & 0.5333 & 0.4722 & \textbf{0.5889} & 0.5333 \\
    SpokenArabicDigits & \underline{0.9850} & \textbf{0.9914} & 0.9759 & 0.8425 & 0.9432 & 0.9031 & 0.9527 & 0.7849 & 0.9704 \\
    StandWalkJump & 0.2667 & \textbf{0.5333} & \underline{0.4667} & 0.4400 & 0.4000 & 0.3333 & 0.3333 & 0.3333 & \textbf{0.5333} \\
    UWaveGestureLibrary & 0.8812 & \underline{0.9031} & 0.8500 & 0.8250 & 0.7781 & 0.7375 & 0.8875 & 0.7937 & \textbf{0.9125} \\
    \midrule
    \textbf{Avg. Acc. (\%)} & 65.33 & 67.96 & 68.17 & 69.34 & \underline{70.50} & 69.54 & 69.72 & 57.92 & \textbf{72.27} \\
    \textbf{Avg. Rank} & 5.18 & 4.37 & 5.48 & 5.22 & \underline{4.02} & 4.63 & 4.92 & 7.62 & \textbf{3.57} \\
    \bottomrule
  \end{tabular}%
  }
\end{table}


\begin{center}
\captionsetup{font=footnotesize}
\captionof{table}{Per-dataset classification accuracy on the complete UCR-128 archive. Bold and underlined values denote the best and second-best results per dataset. Method headers are abbreviated: Chorus=ChorusTIC, MOM=MOMENT+SVM, Mantis=Mantis+RF, MV2-RF=MantisV2+RF, MV2-LR=MantisV2+LR, Uni=UniShape+RF, and NuTime=NuTime+RF.}
\label{tab:ucr_full_results_11methods}
\end{center}
\vspace{-0.7em}
\begingroup
\fontsize{5.1pt}{5.45pt}\selectfont
\setlength{\tabcolsep}{0pt}
\renewcommand{\arraystretch}{1.0}
\noindent\begin{tabular*}{\textwidth}{@{\extracolsep{\fill}}l*{11}{r}@{}}
\toprule
Dataset & MOM & Mantis & MV2-RF & MV2-LR & Uni & NuTime & TabICL & TabICLv2 & TiCT & TIC-FM & Chorus \\
\midrule
Earthquakes & 0.7266 & \underline{0.7482} & \underline{0.7482} & 0.7122 & 0.7338 & 0.7266 & \underline{0.7482} & 0.7410 & \textbf{0.8261} & \underline{0.7482} & \underline{0.7482} \\
WordSynonyms & 0.6160 & 0.5448 & 0.5386 & 0.6160 & 0.5580 & 0.3966 & 0.5956 & \textbf{0.6301} & 0.5870 & \underline{0.6176} & \underline{0.6176} \\
ShapeletSim & \underline{0.9722} & 0.9200 & 0.8267 & 0.8222 & \textbf{1.0000} & 0.6333 & 0.4889 & 0.5333 & 0.9000 & 0.7944 & 0.9278 \\
ProximalPhalanxOutlineAgeGroup & 0.8488 & \textbf{0.8576} & 0.8449 & 0.8293 & 0.8439 & 0.8195 & 0.8390 & 0.8439 & \underline{0.8525} & 0.8341 & 0.8439 \\
PigAirwayPressure & 0.0817 & 0.4606 & \underline{0.7894} & \textbf{0.8510} & 0.1779 & 0.2500 & 0.1635 & 0.1731 & 0.1538 & 0.7067 & 0.4856 \\
PigArtPressure & 0.8606 & 0.8885 & \underline{0.9452} & \textbf{0.9519} & 0.8846 & 0.7019 & 0.2500 & 0.2356 & 0.4808 & 0.9375 & 0.8702 \\
ShapesAll & \textbf{0.8733} & 0.8140 & 0.8463 & \textbf{0.8733} & 0.8117 & 0.7050 & 0.7917 & 0.8000 & 0.7583 & 0.8483 & \underline{0.8517} \\
SmallKitchenAppliances & 0.7440 & 0.8112 & 0.7856 & 0.8027 & \underline{0.8267} & 0.7947 & 0.7653 & 0.7013 & 0.7600 & 0.8213 & \textbf{0.8347} \\
ECG5000 & 0.9447 & 0.9213 & 0.9376 & 0.9231 & 0.9353 & 0.8776 & \underline{0.9449} & \textbf{0.9458} & 0.9440 & 0.9376 & 0.9391 \\
HandOutlines & 0.9216 & \underline{0.9249} & 0.8389 & 0.8946 & 0.9108 & 0.7324 & 0.9243 & \textbf{0.9378} & 0.8623 & 0.8811 & 0.8892 \\
Wafer & \underline{0.9974} & 0.9903 & 0.9907 & 0.9932 & \textbf{0.9995} & 0.9929 & 0.9969 & 0.9958 & 0.9958 & 0.9916 & 0.9950 \\
GunPointAgeSpan & 0.9842 & 0.9911 & 0.9842 & \textbf{1.0000} & 0.9684 & 0.9589 & 0.9810 & \underline{0.9937} & \textbf{1.0000} & 0.9842 & 0.9905 \\
AllGestureWiimoteZ & 0.6157 & 0.6649 & 0.6631 & 0.6514 & 0.6171 & 0.5129 & 0.4486 & 0.5914 & 0.6500 & \underline{0.6686} & \textbf{0.6929} \\
Car & 0.7500 & \underline{0.7867} & 0.7767 & 0.7833 & 0.7833 & 0.6500 & \textbf{0.8333} & \textbf{0.8333} & \textbf{0.8333} & \textbf{0.8333} & 0.6833 \\
FacesUCR & 0.8546 & 0.8245 & 0.8180 & 0.8761 & 0.7790 & 0.5273 & 0.8776 & \underline{0.8966} & \textbf{0.9031} & 0.8893 & 0.8302 \\
Yoga & 0.7300 & 0.8099 & 0.8203 & 0.8277 & 0.8530 & 0.6697 & 0.8677 & \underline{0.8837} & \textbf{0.9424} & 0.8527 & 0.8617 \\
GestureMidAirD1 & 0.6538 & 0.6462 & 0.6677 & \textbf{0.7308} & 0.6923 & 0.5385 & 0.6846 & 0.6923 & 0.6923 & 0.6846 & \underline{0.7154} \\
SonyAIBORobotSurface1 & 0.8968 & 0.7704 & 0.7524 & 0.8153 & 0.8220 & 0.6722 & 0.6705 & \underline{0.9135} & \textbf{0.9839} & 0.7188 & 0.7770 \\
ToeSegmentation2 & \textbf{0.9231} & \underline{0.9200} & 0.8723 & 0.9154 & 0.8538 & 0.8154 & 0.7923 & 0.7846 & 0.8125 & 0.9077 & 0.8923 \\
MiddlePhalanxOutlineAgeGroup & 0.6299 & 0.5870 & 0.5948 & 0.5519 & 0.5779 & 0.5974 & 0.6299 & 0.6299 & \textbf{0.7143} & 0.6364 & \underline{0.6494} \\
ProximalPhalanxTW & 0.8098 & 0.7659 & 0.7893 & 0.7854 & 0.8000 & 0.5854 & \textbf{0.8293} & \underline{0.8244} & 0.8033 & 0.7902 & 0.8049 \\
NonInvasiveFetalECGThorax2 & 0.9191 & 0.6778 & 0.8495 & 0.8997 & 0.8707 & 0.6992 & \underline{0.9410} & \textbf{0.9659} & 0.8590 & 0.8718 & 0.9084 \\
CricketY & 0.7308 & 0.7374 & 0.7508 & 0.7385 & 0.7436 & 0.5769 & 0.6385 & 0.6769 & 0.6944 & \textbf{0.7897} & \underline{0.7872} \\
GunPointMaleVersusFemale & 0.9842 & \underline{0.9968} & \underline{0.9968} & \underline{0.9968} & \underline{0.9968} & 0.9810 & \textbf{1.0000} & \textbf{1.0000} & \textbf{1.0000} & \underline{0.9968} & \underline{0.9968} \\
DodgerLoopDay & 0.4000 & 0.4975 & 0.4700 & 0.5125 & 0.5375 & 0.4500 & \underline{0.6375} & 0.6250 & \textbf{0.7143} & 0.4750 & 0.6000 \\
Worms & \textbf{0.7792} & 0.6260 & 0.6987 & 0.7403 & \underline{0.7532} & 0.7143 & 0.5584 & 0.5844 & 0.7200 & 0.6753 & \underline{0.7532} \\
EthanolLevel & 0.3680 & 0.2980 & 0.3376 & 0.4000 & 0.2960 & 0.3480 & \underline{0.6620} & \textbf{0.8040} & 0.3800 & 0.3100 & 0.4340 \\
TwoLeadECG & 0.9956 & \underline{0.9961} & 0.9914 & \textbf{0.9974} & 0.9245 & 0.8534 & 0.9157 & 0.9271 & 0.9655 & 0.9903 & 0.9605 \\
ECGFiveDays & \underline{0.9721} & 0.8997 & 0.8978 & 0.9605 & 0.7944 & 0.6992 & \textbf{0.9803} & 0.9617 & 0.9432 & 0.9617 & 0.8746 \\
DodgerLoopGame & 0.8406 & 0.7246 & 0.6420 & 0.6232 & 0.7319 & 0.6449 & 0.7609 & 0.8188 & \textbf{0.9375} & 0.5942 & \underline{0.8841} \\
Trace & \textbf{1.0000} & \textbf{1.0000} & \textbf{1.0000} & \textbf{1.0000} & \textbf{1.0000} & \underline{0.9900} & 0.9600 & 0.9600 & \textbf{1.0000} & \textbf{1.0000} & \textbf{1.0000} \\
Plane & \textbf{1.0000} & \textbf{1.0000} & \textbf{1.0000} & \textbf{1.0000} & \textbf{1.0000} & \underline{0.9905} & \underline{0.9905} & \textbf{1.0000} & \textbf{1.0000} & \textbf{1.0000} & \textbf{1.0000} \\
Meat & 0.8333 & \underline{0.9333} & 0.9000 & \textbf{0.9500} & \underline{0.9333} & 0.8167 & \textbf{0.9500} & \underline{0.9333} & 0.6667 & 0.9000 & 0.8833 \\
Crop & 0.6994 & 0.6689 & 0.6971 & 0.7035 & 0.7155 & 0.6018 & \underline{0.8134} & \textbf{0.8214} & 0.7242 & 0.6514 & 0.7433 \\
UWaveGestureLibraryZ & 0.7398 & 0.7225 & 0.7061 & 0.7272 & 0.7370 & 0.7256 & 0.7404 & 0.7524 & \textbf{0.7790} & 0.7661 & \underline{0.7694} \\
SemgHandMovementCh2 & 0.4200 & 0.7209 & 0.6378 & 0.6000 & \underline{0.7689} & 0.6422 & 0.5800 & 0.6578 & 0.6111 & 0.7489 & \textbf{0.7933} \\
CricketX & 0.7154 & 0.7328 & 0.7405 & 0.7256 & 0.7359 & 0.6077 & 0.6513 & 0.6590 & 0.6528 & \textbf{0.8103} & \underline{0.7590} \\
DodgerLoopWeekend & 0.9565 & 0.9536 & 0.9551 & 0.9565 & 0.9638 & 0.8841 & 0.9783 & \underline{0.9855} & \textbf{1.0000} & 0.9420 & \underline{0.9855} \\
SmoothSubspace & 0.9667 & 0.9080 & 0.9040 & 0.9467 & 0.9267 & 0.9000 & \underline{0.9933} & \textbf{1.0000} & 0.9333 & 0.9333 & 0.9867 \\
PigCVP & 0.7981 & 0.7644 & \underline{0.9000} & \textbf{0.9135} & 0.4375 & 0.6058 & 0.2260 & 0.2019 & 0.2115 & 0.8750 & 0.8173 \\
DistalPhalanxTW & 0.6691 & 0.6820 & 0.6835 & 0.6691 & 0.6331 & 0.6259 & 0.6906 & \textbf{0.7050} & \underline{0.6909} & 0.6835 & 0.6906 \\
Mallat & 0.8768 & 0.8829 & 0.8829 & 0.9258 & 0.8840 & 0.7821 & 0.9531 & \underline{0.9727} & \textbf{0.9875} & 0.9326 & 0.9561 \\
GunPoint & \textbf{0.9933} & 0.9693 & 0.9853 & \underline{0.9867} & 0.9733 & 0.8867 & 0.9533 & 0.9667 & 0.9500 & \textbf{0.9933} & \underline{0.9867} \\
MelbournePedestrian & 0.8421 & 0.8999 & 0.9398 & 0.9512 & 0.9377 & 0.8827 & \underline{0.9799} & \textbf{0.9836} & 0.8579 & 0.9582 & 0.9471 \\
FordB & 0.7938 & 0.7341 & 0.7849 & \underline{0.8074} & 0.7840 & 0.6802 & 0.7037 & 0.7185 & \textbf{0.8761} & 0.7568 & 0.7321 \\
InsectEPGRegularTrain & \textbf{1.0000} & \textbf{1.0000} & \textbf{1.0000} & \textbf{1.0000} & \textbf{1.0000} & \textbf{1.0000} & \textbf{1.0000} & \textbf{1.0000} & \textbf{1.0000} & \textbf{1.0000} & \textbf{1.0000} \\
FreezerSmallTrain & 0.8077 & 0.8022 & 0.8312 & 0.8632 & 0.9302 & \underline{0.9526} & 0.7828 & 0.8958 & \textbf{0.9618} & 0.8568 & 0.9242 \\
Phoneme & 0.2764 & \underline{0.3270} & 0.3155 & 0.3128 & \textbf{0.3312} & 0.1609 & 0.1329 & 0.1361 & 0.1934 & 0.3149 & 0.2801 \\
GunPointOldVersusYoung & 0.9492 & 0.9968 & \underline{0.9987} & \textbf{1.0000} & 0.9968 & \textbf{1.0000} & \textbf{1.0000} & \textbf{1.0000} & \textbf{1.0000} & \textbf{1.0000} & 0.9968 \\
Lightning2 & 0.7541 & 0.8033 & 0.7279 & 0.8033 & 0.7705 & 0.7213 & 0.7213 & 0.7213 & \underline{0.8333} & \textbf{0.8525} & 0.7541 \\
BirdChicken & 0.9000 & \underline{0.9900} & 0.8000 & 0.9000 & 0.9000 & 0.7500 & 0.7500 & 0.6000 & \textbf{1.0000} & 0.7500 & \textbf{1.0000} \\
Lightning7 & 0.6849 & \underline{0.7534} & 0.6219 & 0.7397 & 0.7123 & 0.5890 & 0.7260 & \textbf{0.7671} & 0.4667 & \underline{0.7534} & 0.6712 \\
CBF & 0.9767 & 0.9889 & 0.9840 & 0.9967 & 0.9933 & 0.9267 & 0.9211 & 0.9389 & \textbf{1.0000} & \underline{0.9989} & 0.9978 \\
MiddlePhalanxTW & 0.5974 & 0.5260 & 0.5156 & 0.5260 & 0.5519 & 0.4351 & \textbf{0.6234} & \underline{0.6169} & 0.5273 & 0.5714 & 0.5714 \\
SemgHandGenderCh2 & 0.7617 & 0.8937 & 0.8353 & 0.8100 & 0.8983 & 0.8467 & 0.8717 & \textbf{0.9717} & 0.8889 & 0.9183 & \underline{0.9250} \\
SemgHandSubjectCh2 & 0.6556 & 0.8031 & 0.7436 & 0.7489 & 0.8556 & 0.6689 & 0.8356 & \textbf{0.9244} & \underline{0.9000} & 0.8400 & 0.8378 \\
Rock & 0.7400 & 0.7640 & 0.7200 & \underline{0.8400} & 0.6400 & 0.5600 & 0.6200 & 0.7200 & \textbf{0.8571} & 0.6400 & 0.8200 \\
EOGVerticalSignal & \textbf{0.5166} & 0.4575 & 0.4387 & 0.4751 & 0.4724 & 0.2597 & 0.4254 & 0.5000 & \underline{0.5139} & 0.4834 & 0.4669 \\
InlineSkate & 0.3236 & 0.3535 & 0.3775 & 0.3873 & 0.3255 & 0.2873 & 0.3709 & 0.3545 & \textbf{0.5231} & \underline{0.4273} & 0.4236 \\
ScreenType & 0.5360 & 0.4464 & 0.4549 & 0.5093 & 0.5147 & 0.4427 & 0.4240 & 0.4400 & \textbf{0.7200} & 0.5173 & \underline{0.5573} \\
ProximalPhalanxOutlineCorrect & 0.7595 & 0.8131 & 0.8323 & \underline{0.8557} & 0.8488 & 0.7904 & \textbf{0.9210} & \textbf{0.9210} & 0.8202 & 0.7801 & 0.8351 \\
GesturePebbleZ1 & 0.9128 & \textbf{0.9279} & 0.9058 & 0.9128 & 0.9186 & 0.8721 & 0.8488 & 0.8837 & 0.8966 & 0.9070 & \underline{0.9244} \\
SwedishLeaf & 0.9376 & 0.9274 & 0.9248 & \textbf{0.9504} & 0.9136 & 0.8560 & \textbf{0.9504} & \underline{0.9440} & 0.9000 & 0.9296 & 0.9392 \\
RefrigerationDevices & 0.4987 & 0.5019 & 0.5184 & 0.5093 & \textbf{0.5867} & \underline{0.5440} & 0.4720 & 0.4987 & \textbf{0.5867} & 0.5387 & 0.5093 \\
Beef & 0.6000 & 0.6533 & 0.5467 & 0.6000 & 0.6667 & 0.4000 & \underline{0.7333} & \textbf{0.8000} & 0.4000 & \underline{0.7333} & 0.6667 \\
Strawberry & 0.9216 & 0.9503 & 0.9535 & 0.9432 & 0.9297 & 0.8649 & \underline{0.9838} & \textbf{0.9865} & 0.9286 & 0.9405 & 0.9622 \\
Symbols & \underline{0.9678} & 0.9574 & 0.8880 & 0.9276 & 0.9166 & 0.8101 & 0.9126 & 0.9176 & \textbf{1.0000} & 0.9598 & 0.9508 \\
Herring & 0.5938 & 0.6375 & \textbf{0.7031} & 0.6875 & 0.5938 & 0.4531 & 0.6094 & 0.5469 & \underline{0.6923} & 0.6250 & 0.6406 \\
UMD & 0.9792 & 0.9694 & 0.9778 & \underline{0.9931} & 0.9861 & 0.7639 & 0.9306 & \underline{0.9931} & \textbf{1.0000} & \underline{0.9931} & \underline{0.9931} \\
InsectWingbeatSound & 0.5980 & 0.5127 & 0.4976 & 0.4960 & 0.5318 & 0.4308 & \underline{0.6505} & \textbf{0.6682} & 0.6409 & 0.5162 & 0.5091 \\
MiddlePhalanxOutlineCorrect & 0.6529 & 0.8055 & 0.8296 & \underline{0.8488} & 0.8144 & 0.7423 & \textbf{0.8522} & 0.8247 & 0.7528 & 0.8144 & 0.8282 \\
ElectricDevices & 0.7513 & 0.7228 & 0.6946 & 0.6932 & \underline{0.7548} & 0.6964 & 0.6623 & 0.6933 & \textbf{0.8065} & 0.7199 & 0.7426 \\
OSULeaf & 0.8926 & 0.8636 & 0.8727 & \underline{0.9215} & 0.8182 & 0.7314 & 0.6116 & 0.5909 & \textbf{0.9333} & 0.8430 & 0.8802 \\
GesturePebbleZ2 & 0.9114 & \textbf{0.9241} & 0.9051 & \underline{0.9177} & 0.8797 & 0.8101 & 0.7658 & 0.7468 & 0.8966 & 0.8228 & 0.8165 \\
StarLightCurves & 0.9733 & 0.9761 & 0.9792 & 0.9741 & 0.9779 & 0.9757 & 0.9709 & 0.9754 & 0.9773 & \underline{0.9796} & \textbf{0.9809} \\
AllGestureWiimoteX & \underline{0.7129} & 0.6609 & 0.6297 & 0.6900 & 0.6129 & 0.5571 & 0.4929 & 0.6300 & 0.7000 & 0.7000 & \textbf{0.7514} \\
Adiac & 0.2916 & 0.7253 & 0.8046 & 0.8005 & 0.7749 & 0.5729 & 0.8005 & \textbf{0.8465} & 0.5714 & 0.6573 & \underline{0.8184} \\
GestureMidAirD3 & 0.3538 & 0.3400 & 0.3723 & 0.4077 & 0.3769 & 0.3538 & 0.3308 & 0.4154 & \textbf{0.4615} & 0.3692 & \underline{0.4308} \\
FordA & \textbf{0.9417} & 0.8565 & 0.9038 & \underline{0.9295} & 0.8932 & 0.8659 & 0.8727 & 0.9068 & 0.8760 & 0.8962 & 0.8697 \\
FreezerRegularTrain & 0.9488 & 0.9354 & 0.9537 & 0.9842 & 0.9814 & 0.9737 & 0.9895 & \textbf{0.9979} & 0.9600 & 0.9811 & \underline{0.9951} \\
InsectEPGSmallTrain & 0.9558 & \textbf{1.0000} & \underline{0.9976} & \textbf{1.0000} & 0.9960 & 0.9719 & \textbf{1.0000} & \textbf{1.0000} & \textbf{1.0000} & 0.9799 & \textbf{1.0000} \\
MixedShapesSmallTrain & 0.8874 & 0.8884 & 0.8837 & 0.8953 & \underline{0.9142} & 0.8309 & 0.8722 & 0.8561 & \textbf{0.9643} & 0.9122 & 0.9118 \\
DistalPhalanxOutlineAgeGroup & 0.7554 & 0.7885 & 0.7669 & 0.7410 & \underline{0.7986} & 0.7194 & 0.7626 & 0.7410 & \textbf{0.8679} & 0.7410 & 0.7626 \\
PLAID & 0.7523 & 0.8086 & 0.7940 & 0.7765 & \underline{0.8287} & 0.6089 & 0.5773 & 0.8007 & 0.6818 & 0.7635 & \textbf{0.8641} \\
Fish & 0.8629 & \textbf{0.9383} & 0.9086 & \underline{0.9200} & 0.8743 & 0.8286 & 0.8743 & 0.8857 & 0.5429 & 0.8743 & 0.9143 \\
DistalPhalanxOutlineCorrect & 0.7862 & 0.7543 & 0.7761 & 0.7645 & 0.7862 & 0.7391 & 0.7790 & \underline{0.7971} & \textbf{0.8295} & 0.7717 & 0.7935 \\
MoteStrain & 0.8986 & 0.9059 & \underline{0.9313} & 0.9241 & 0.8674 & 0.9137 & 0.8850 & 0.8922 & \textbf{0.9685} & 0.9185 & 0.8474 \\
Ham & 0.7048 & 0.6743 & 0.6629 & 0.6286 & 0.5905 & 0.6095 & \underline{0.7143} & \textbf{0.7333} & \underline{0.7143} & 0.6571 & 0.6667 \\
FaceAll & \underline{0.8077} & 0.7807 & 0.7146 & 0.7414 & 0.7870 & 0.5680 & 0.7751 & 0.7775 & \textbf{0.9383} & 0.7444 & 0.7036 \\
FiftyWords & \textbf{0.7736} & 0.6295 & 0.5974 & 0.6835 & 0.6857 & 0.4440 & 0.7165 & \underline{0.7451} & 0.6087 & 0.6923 & 0.7231 \\
MixedShapesRegularTrain & 0.9460 & 0.9391 & 0.9292 & 0.9410 & 0.9452 & 0.9175 & 0.9320 & 0.9365 & \textbf{0.9623} & 0.9443 & \underline{0.9513} \\
Fungi & \textbf{1.0000} & 0.8022 & 0.8172 & 0.8763 & 0.7312 & 0.3548 & 0.8065 & 0.8978 & \underline{0.9545} & 0.7849 & 0.5430 \\
EOGHorizontalSignal & 0.5552 & 0.5917 & 0.5845 & 0.5773 & 0.5442 & 0.4116 & 0.4917 & 0.5387 & \textbf{0.7222} & 0.5801 & \underline{0.6022} \\
UWaveGestureLibraryY & 0.7281 & 0.6747 & 0.6682 & 0.7108 & 0.7362 & 0.6834 & 0.7203 & 0.7233 & \textbf{0.7902} & 0.7426 & \underline{0.7610} \\
Computers & 0.7280 & 0.7288 & 0.7256 & 0.7000 & \underline{0.7800} & 0.7720 & 0.6280 & 0.6360 & 0.6800 & 0.7480 & \textbf{0.8080} \\
PickupGestureWiimoteZ & 0.7200 & 0.7920 & 0.7840 & 0.7600 & \underline{0.8000} & 0.6600 & 0.7000 & 0.7000 & 0.5000 & 0.7200 & \textbf{0.9200} \\
CricketZ & 0.7179 & 0.7733 & \underline{0.7841} & 0.7513 & 0.7667 & 0.6026 & 0.6615 & 0.7026 & 0.7500 & \textbf{0.8077} & \textbf{0.8077} \\
ACSF1 & 0.6800 & 0.7820 & 0.7680 & 0.6900 & 0.7400 & 0.4400 & \underline{0.8200} & \textbf{0.8600} & 0.3500 & 0.6700 & 0.7400 \\
UWaveGestureLibraryX & 0.7954 & 0.7614 & 0.7319 & 0.7761 & 0.8057 & 0.7806 & 0.7959 & 0.8026 & \textbf{0.8348} & 0.8132 & \underline{0.8303} \\
ArrowHead & 0.6971 & 0.7166 & 0.7886 & \underline{0.8000} & 0.7943 & 0.5600 & 0.7371 & 0.7886 & \textbf{0.8500} & 0.7771 & 0.7829 \\
\bottomrule
\end{tabular*}
\endgroup

\clearpage
\begin{center}
\captionsetup{font=footnotesize}
\captionof*{table}{Table~\ref{tab:ucr_full_results_11methods}: Per-dataset classification accuracy on UCR-128 (continued).}
\end{center}
\vspace{-0.7em}
\begingroup
\fontsize{5.1pt}{5.45pt}\selectfont
\setlength{\tabcolsep}{0pt}
\renewcommand{\arraystretch}{1.0}
\noindent\begin{tabular*}{\textwidth}{@{\extracolsep{\fill}}l*{11}{r}@{}}
\toprule
Dataset & MOM & Mantis & MV2-RF & MV2-LR & Uni & NuTime & TabICL & TabICLv2 & TiCT & TIC-FM & Chorus \\
\midrule
SyntheticControl & 0.9633 & 0.9753 & 0.9753 & 0.9733 & 0.9633 & 0.9600 & 0.9833 & \underline{0.9867} & 0.9667 & \textbf{0.9933} & \textbf{0.9933} \\
TwoPatterns & \underline{0.9838} & 0.8708 & 0.9274 & 0.9758 & 0.8750 & 0.8085 & 0.9002 & 0.9355 & 0.8040 & \textbf{0.9852} & 0.9810 \\
ECG200 & 0.8700 & 0.8220 & 0.8620 & 0.8600 & 0.8700 & 0.8100 & \textbf{0.8900} & \underline{0.8800} & 0.7000 & 0.8200 & 0.8500 \\
Coffee & 0.8929 & 0.9571 & \textbf{1.0000} & \textbf{1.0000} & \underline{0.9643} & 0.8571 & \textbf{1.0000} & \textbf{1.0000} & \textbf{1.0000} & \textbf{1.0000} & \underline{0.9643} \\
Wine & 0.5000 & 0.7667 & 0.7444 & 0.6667 & 0.6296 & 0.5926 & 0.7037 & 0.7037 & 0.7273 & \textbf{0.8704} & \underline{0.8148} \\
PowerCons & 0.9000 & 0.9144 & 0.9289 & 0.9389 & 0.9389 & 0.9111 & \underline{0.9889} & \textbf{1.0000} & 0.8889 & 0.9611 & 0.9722 \\
UWaveGestureLibraryAll & 0.9227 & 0.8382 & 0.7951 & 0.8495 & 0.8755 & 0.8023 & \underline{0.9651} & \textbf{0.9698} & 0.9576 & 0.8889 & 0.8814 \\
PhalangesOutlinesCorrect & 0.7016 & 0.7699 & 0.8214 & 0.8042 & 0.7995 & 0.7075 & \underline{0.8578} & \textbf{0.8601} & 0.8346 & 0.7506 & 0.7867 \\
ItalyPowerDemand & 0.9504 & 0.9044 & 0.9158 & 0.9125 & 0.7940 & 0.8562 & \underline{0.9621} & \textbf{0.9689} & 0.9273 & 0.9174 & 0.8678 \\
MedicalImages & 0.7618 & 0.6966 & 0.7313 & 0.7526 & 0.7184 & 0.5934 & \underline{0.7987} & \textbf{0.8342} & 0.7965 & 0.7829 & 0.7803 \\
NonInvasiveFetalECGThorax1 & 0.9033 & 0.6159 & 0.8117 & 0.8850 & 0.8417 & 0.6738 & \underline{0.9237} & \textbf{0.9562} & 0.8457 & 0.8504 & 0.9003 \\
Haptics & 0.4968 & 0.4721 & 0.4877 & 0.4643 & \underline{0.5130} & 0.3994 & 0.4513 & 0.4675 & \textbf{0.6304} & 0.4481 & \underline{0.5130} \\
ChlorineConcentration & 0.5716 & 0.6765 & 0.6700 & 0.6609 & 0.6703 & 0.3732 & \underline{0.9703} & \textbf{0.9927} & 0.7471 & 0.5656 & 0.6771 \\
HouseTwenty & 0.9580 & 0.9445 & 0.9395 & 0.9580 & 0.9496 & 0.9160 & 0.7395 & 0.8319 & \textbf{1.0000} & 0.9496 & \underline{0.9748} \\
BME & 0.9800 & 0.9347 & 0.7707 & 0.8067 & 0.9467 & 0.6400 & 0.9800 & \textbf{1.0000} & 0.6667 & 0.9533 & \underline{0.9933} \\
CinCECGTorso & 0.7565 & 0.6584 & 0.7209 & 0.7572 & 0.7942 & 0.6283 & 0.8080 & \underline{0.8783} & \textbf{1.0000} & 0.6964 & 0.6688 \\
GestureMidAirD2 & 0.5231 & \underline{0.6138} & 0.5662 & 0.5769 & 0.6077 & 0.4538 & 0.5846 & \textbf{0.6154} & 0.5769 & \textbf{0.6154} & 0.5846 \\
SonyAIBORobotSurface2 & \underline{0.9570} & 0.8306 & 0.9142 & 0.9265 & 0.8206 & 0.7702 & 0.8300 & 0.8174 & \textbf{0.9796} & 0.8804 & 0.9035 \\
DiatomSizeReduction & 0.7092 & 0.8575 & 0.8085 & 0.8824 & 0.8464 & 0.8105 & 0.9477 & \underline{0.9771} & \textbf{1.0000} & 0.9542 & 0.9444 \\
WormsTwoClass & \underline{0.8052} & 0.7922 & 0.7506 & 0.7662 & 0.7922 & 0.7532 & 0.5974 & 0.6234 & \textbf{0.8462} & 0.7662 & 0.7403 \\
ToeSegmentation1 & 0.9386 & \underline{0.9649} & \textbf{0.9711} & \underline{0.9649} & 0.9123 & 0.8333 & 0.6535 & 0.6053 & 0.9630 & 0.9035 & 0.9079 \\
Chinatown & 0.9650 & 0.8426 & 0.9364 & 0.9446 & 0.9592 & 0.6210 & \underline{0.9796} & \textbf{0.9883} & 0.9444 & 0.9738 & 0.9621 \\
OliveOil & 0.4000 & \underline{0.9133} & 0.8467 & 0.4000 & 0.7333 & 0.5667 & 0.9000 & \textbf{0.9333} & 0.3333 & 0.6333 & 0.8333 \\
BeetleFly & \textbf{0.9500} & 0.8300 & \underline{0.9400} & 0.8500 & \textbf{0.9500} & 0.7500 & 0.9000 & 0.7000 & 0.5000 & 0.9000 & 0.7000 \\
AllGestureWiimoteY & 0.7443 & 0.6483 & 0.6526 & 0.6843 & 0.6543 & 0.5371 & 0.4957 & 0.6929 & \textbf{0.7600} & 0.7243 & \underline{0.7457} \\
ShakeGestureWiimoteZ & 0.9000 & 0.8840 & \underline{0.9080} & 0.8800 & 0.8400 & 0.8200 & 0.7600 & 0.7400 & 0.9000 & \textbf{0.9400} & \textbf{0.9400} \\
FaceFour & 0.7955 & 0.9455 & 0.9227 & \underline{0.9545} & 0.8864 & 0.7841 & 0.8750 & 0.8864 & \textbf{1.0000} & 0.9205 & 0.8295 \\
LargeKitchenAppliances & \textbf{0.8480} & 0.7904 & 0.6795 & 0.7920 & \underline{0.8427} & 0.7653 & 0.7093 & 0.6720 & 0.7333 & 0.7920 & 0.8267 \\
\midrule
\textbf{Avg. Acc. (\%)} & 77.98 & 78.67 & 78.79 & \underline{80.03} & 78.86 & 69.39 & 76.83 & 78.88 & 79.17 & 80.01 & \textbf{81.16} \\
\textbf{Avg. Rank} & 6.11 & 6.42 & 6.51 & 5.50 & 5.83 & 9.55 & 6.38 & 5.15 & \underline{4.81} & 5.32 & \textbf{4.43} \\
\bottomrule
\end{tabular*}
\endgroup



\end{document}